\documentclass[10pt,twocolumn,letterpaper]{article}

\usepackage{cvpr}
\usepackage{times}
\usepackage{url}
\usepackage[numbers]{natbib}
\usepackage[pdftex]{graphicx}
\usepackage{subcaption}
\usepackage{amsmath}
\usepackage{amssymb}
\usepackage{wrapfig}
\usepackage{xr}
\usepackage{algorithm}
\usepackage{algpseudocode}

\usepackage[pagebackref=true,breaklinks=true,letterpaper=true,colorlinks,bookmarks=false]{hyperref}

\newcommand{\comment}[1]{}

\newcommand{\angs}{{\r{A}}}
\newcommand{\R}{\mathbb{R}}
\newcommand{\SO}{\mathcal{SO}(3)}
\newcommand{\eye}{\mathbf{I}}

\newcommand{\projdir}{\mathbf{R}}
\newcommand{\proj}[1]{ \mathbf{P}_{ { #1 } } }
\newcommand{\ftproj}[1]{ \tilde{\mathbf{P}}_{ { #1 } } }

\newcommand{\ctfparam}{\theta}
\newcommand{\ctf}[1]{\mathbf{C}_{#1}}
\newcommand{\ftctf}[1]{\tilde{\mathbf{C}}_{#1}}

\newcommand{\shiftdir}{\mathbf{t}}
\newcommand{\shift}[1]{ \mathbf{S}_{ { #1 } } }
\newcommand{\ftshift}[1]{ \tilde{\mathbf{S}}_{ { #1 } } }

\newcommand{\img}{\mathcal{I}}
\newcommand{\ftimg}{\tilde{\mathcal{I}}}
\newcommand{\density}{\mathcal{V}}
\newcommand{\ftdensity}{\tilde{\density}}
\newcommand{\densitygrad}{d\density}
\newcommand{\rad}{\omega} 

\newcommand{\iter}{{\tau}}
\newcommand{\obj}{f}
\newcommand{\dobj}{\mathbf{g}}
\newcommand{\isp}{q}
\newcommand{\isP}{\mathbf{q}}
\newcommand{\kernmat}{\mathbf{K}}

\newcommand{\N}{D} 

\newcommand{\noiseStd}{\sigma}

\newcommand{\data}{\mathfrak{D}}

\cvprfinalcopy 

\def\cvprPaperID{1361} 

\ifcvprfinal\fi

\begin{document}
\title{Building Proteins in a Day: Efficient 3D Molecular Reconstruction}

\author{Marcus A. Brubaker\\
\and
Ali Punjani\\
University of Toronto\\
{\tt\small \{mbrubake,alipunjani,fleet\}@cs.toronto.edu}
\and
David J. Fleet\\
}

\maketitle

\begin{abstract}
Discovering the 3D atomic structure of molecules such as proteins and 
viruses is a fundamental research problem in biology and medicine.
Electron Cryomicroscopy (Cryo-EM) is a promising vision-based technique
for structure estimation which attempts to reconstruct 3D structures from
2D images.
This paper addresses the challenging problem of 3D reconstruction from
2D Cryo-EM images.
A new framework for estimation is introduced which relies on modern stochastic
optimization techniques to scale to large datasets.
We also introduce a novel technique which reduces the cost of evaluating 
the objective function during optimization by over five orders or magnitude.
The net result is an approach capable of estimating 3D molecular structure
from large scale datasets in about a day on a single workstation.
\end{abstract}

\section{Introduction}

Discovering the 3D atomic structure of molecules such as proteins
and viruses is a fundamental research problem in biology and medicine.
The ability to routinely determine the 3D structure of such molecules
would potentially revolutionize the process of drug development and
accelerate research into fundamental biological processes.
Electron Cryomicroscopy (Cryo-EM) is an emerging vision-based approach
to 3D macromolecular structure determination that is applicable to
medium to large-sized molecules in their native state.
This is in contrast to X-ray crystallography which requires a crystal 
of the target molecule, which are often impossible to 
grow \citep{Rupp2009} or nuclear magnetic resonance (NMR) spectroscopy 
which is limited to relatively small molecules \citep{Keeler2010}.

\begin{figure}
	\includegraphics[width=0.48\textwidth]{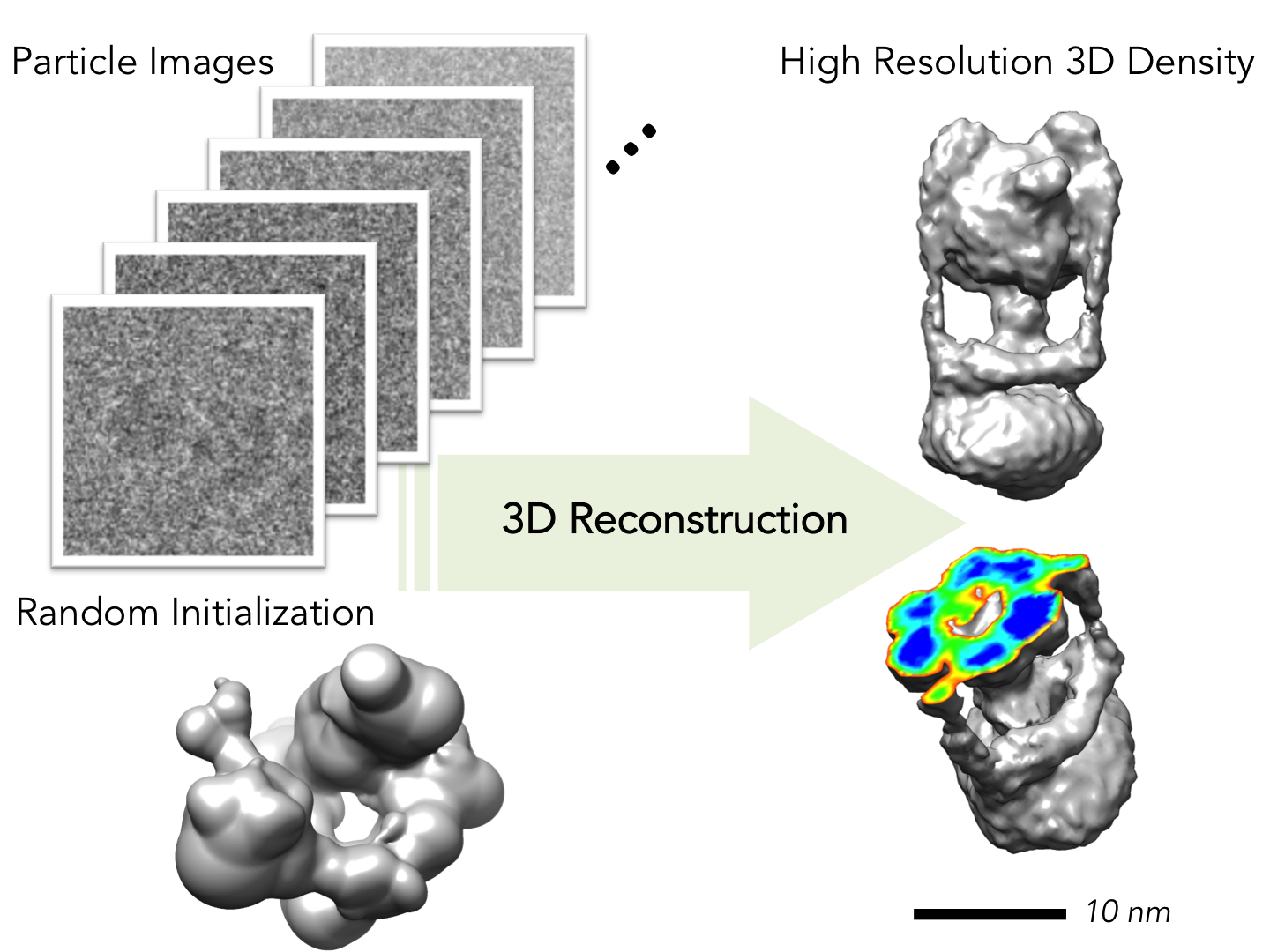}
\caption{\label{fig:kicker}
The goal is to reconstruct the 3D structure of a molecule (right),
at nanometer scales, from a large number of noisy, uncalibrated
2D projections obtained from cryogenically frozen samples in an
electron microscope (left).
}
\end{figure}

The Cryo-EM reconstruction task is to estimate the 3D density 
of a target molecule from a large set of images of the molecule (called
particle images).  The problem is similar in spirit to multi-view scene 
carving \citep{Bhotika2002,Kutulakos2000} and to large-scale, uncalibrated
multi-view reconstruction \citep{Agarwal2009}.
Like multi-view scene carving, the goal is to estimate a dense 3D 
occupancy representation of shape from a set of different views, 
but unlike many approaches to scene carving, we do not assume 
calibrated cameras, since we do not know the relative 3D poses of the 
molecule in different images.
Like uncalibrated, multi-view reconstruction, we aim to use very large 
numbers of uncalibrated views to obtain high fidelity 3D reconstructions,
but the low signal-to-noise levels in Cryo-EM (often as low as 
0.05 \citep{Baxter2009}; see Fig.\ \ref{fig:kicker}) and 
the different image formation model prevent the use of common feature 
matching techniques to establish correspondences.
Computed Tomography (CT) \citep{Hsieh2003,Gregson2012} uses a similar
imaging model (orthographic integral projection), however in CT
the projection direction of each image is known whereas with Cryo-EM
the relative pose of each particle is unknown.

%

Existing Cryo-EM techniques, \eg, \citep{Rosa-Trevin2013,Grigorieff2007,Scheres2012,Tang2007},
suffer from two key problems.
First, without good initialization, they converge to poor or incorrect 
solutions \citep{Henderson2012}, often with little indication that something 
went wrong.
Second, they are extremely slow, which limits the number of particles 
images one can use as input to mitigate the effects of noise; \eg, the 
website of the RELION package \citep{Scheres2012} reports requiring two 
weeks on 300 cores to process a dataset with 200,000 images.

\begin{figure*}
\centering
\includegraphics[width=0.95\textwidth]{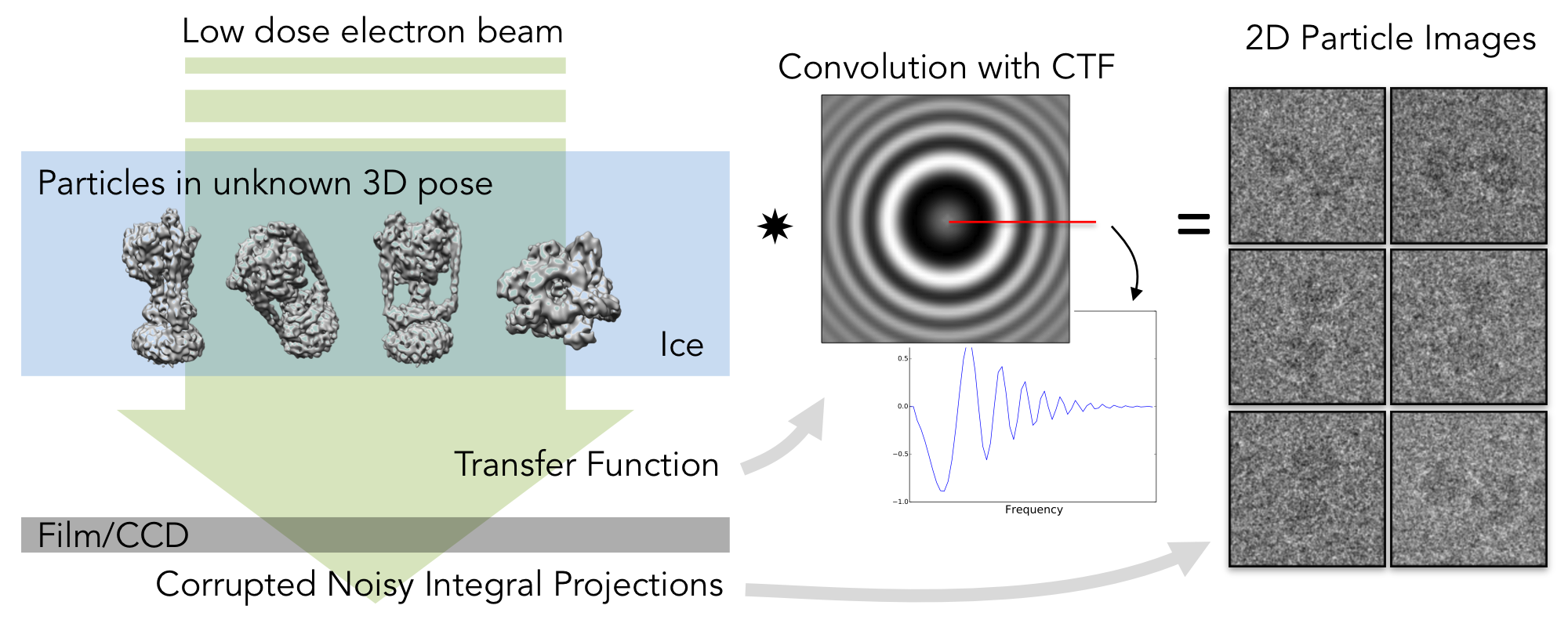}
\caption{
\label{fig:cryoimgs}
A generative image formation model in Cryo-EM.  
The electron beam results in an orthographic integral projection of the
electron density of the specimen.  This projection is modulated by the
Contrast Transfer Function (CTF) and corrupted with noise.
The images pictured here showcase the low SNR typical in Cryo-EM.
The zeros in the CTF (which completely destroy some spatial information)
make estimation particularly challenging, however their locations vary
as a function of microscope parameters. These are set differently across particle 
images in order to mitigate this problem.
Particle images and density from \citep{Lau2012}.
}
\end{figure*}

We introduce a framework for Cryo-EM density estimation, formulating 
the problem as one of stochastic optimization to perform
maximum-a-posteriori (MAP) estimation in a probabilistic model.
The approach is remarkably efficient, providing useful low resolution 
density estimates in an hour.
We also show that our stochastic optimization technique is insensitive 
to initialization, allowing the use of random initializations.
We further introduce a novel importance sampling scheme that dramatically
reduces the computational costs associated with high resolution reconstruction.
This leads to speedups of 100,000-fold or more, allowing structures to be
determined in a day on a modern workstation.
In addition, the proposed framework is flexible, allowing parts of the 
model to be changed and improved without impacting the estimation; \eg, 
we compare the use of three different priors.
To demonstrate our method, we perform reconstructions on two real 
datasets and one synthetic dataset.


\section{Background and Related Work}
\label{sec:background}

In Cryo-EM, a purified solution of the target molecule is cryogenically 
frozen into a thin (single molecule thick) film, and imaged with a 
transmission electron microscope.
A large number of such samples are obtained, each of which provides 
a micrograph containing hundreds of visible, individual molecules.
In a process known as \emph{particle picking}, individual molecules 
are selected, resulting in a stack of cropped {\em particle images}.
Particle picking is often done manually, however there have been 
recent moves to partially or fully automate the process
\citep{Langlois2014,Zhao2013}.
Each particle image provides a noisy view of the molecule, 
but with unknown 3D pose, see Fig.\ \ref{fig:cryoimgs} (right).
The reconstruction task is to estimate the 3D electron density of 
the target molecule from the potentially large set of particle images.

Common approaches to Cryo-EM density estimation, \eg, 
\citep{Rosa-Trevin2013,Grigorieff2007,Tang2007}, use a form
of iterative refinement.
Based on an initial estimate of the 3D density, they determine the 
best matching pose for each particle image.
A new density estimate is then constructed using the Fourier Slice
Theorem (FST); \ie, the 2D Fourier transform of an integral projection
of the density corresponds to a slice through the origin of the 3D
Fourier transform of that density, in a plane perpendicular to
the projection direction \citep{Hsieh2003}.
Using the 3D pose for each particle image, the new density is found
through interpolation and averaging of the observed particle images.

This approach is fundamentally limited in several ways.
Even if one begins with the correct 3D density, the low SNR of particle 
images makes accurately identifying the correct pose for each particle 
nearly impossible. 
This problem is exacerbated when the initial density is inaccurate.
Poor initializations result in estimated structures that are either
clearly wrong (see  Fig.\ \ref{fig:baselines}) or, worse, appear
plausible but are misleading in reality, resulting in incorrectly estimated 
3D structures \citep{Henderson2012}.
Finally, and crucially for the case of density estimation with many particle
images, all data are used at each refinement iteration, causing these methods
to be extremely slow.
\citet{Mallick2006CVPR} proposed an approach which attempted to establish 
weak constraints on the relative 3D poses between different particle images.
This was used to initialize an iterative refinement algorithm to produce 
a final reconstruction.  In contrast, our refinement approach does not 
require an accurate initialization.

To avoid the need to estimate a single 3D pose for each particle
image, Bayesian approaches have been proposed in which the 3D poses 
for the particle images are treated as latent variables, and then
marginalized out numerically.
This approach was originally proposed by \citet{Sigworth1998} for
2D image alignment and later by \citet{Scheres2007} for 3D estimation
and classification. 
It was since been used by \citet{Jaitly2010}, where
batch, gradient-based optimization was performed.
Nevertheless, due to the computational cost of marginalization, the 
method was only applied to small numbers of class-average images
which are produced by clustering, aligning and averaging
individual particle images according to their appearance, to reduce 
noise and the number of particle images used during the optimization.
More recently, pose marginalization was applied directly with particle
images, using a batch Expectation-Maximization algorithm in the RELION
package \citep{Scheres2012}.
However, this approach is extremely computationally expensive.
Here, the proposed approach uses a similar marginalized likelihood, however 
unlike previous methods, stochastic rather than batch optimization is used.
We show that this allows for efficient optimization, and for robustness with
respect to initialization.  
We further introduce a novel importance sampling technique that 
dramatically reduces the computational cost of the marginalization 
when working at high resolutions.

\section{A Framework for 3D Density Estimation}
Here we present our framework for density estimation which includes
a probabilistic generative model of image formation, stochastic optimization
to cope with large-scale datasets, and importance sampling to efficiently 
marginalize over the unknown 3D pose of the particle in each image.

\subsection{Image Formation Model}
In Cryo-EM, particle images are formed as orthographic, integral 
projections of the electron density of a molecule, $\density \in \R^{\N^3}$.
In each image, the density is oriented in an unknown pose,
$\projdir \in \SO$, relative to the direction of the microscope beam.
The projection along this unknown direction is a linear operator, which is 
represented by the matrix $\proj{\projdir} \in \R^{\N^2 \times \N^3}$.
Along with pose, the in-plane translation $\shiftdir \in \R^2$ of each
particle image is unknown, the effect of which is similarly represented by a 
matrix $\shift{\shiftdir} \in \R^{\N^2 \times \N^2}$.
The resulting shifted projection is corrupted by two phenomena: a contrast 
transfer function (CTF) and noise.
The CTF is analogous to the effects of defocus in a conventional light 
camera and can be modelled as a convolution of the projected image.
This linear operation is represented here by the matrix 
$\ctf{\ctfparam} \in \R^{\N^2 \times \N^2}$ where $\ctfparam$ are the
parameters of the CTF model \citep{Reimer2008}.
The Fourier spectrum of a typical CTF is shown in Figure
\ref{fig:cryoimgs}; note the phase changes which result in zero crossings (not typically observed in traditional light cameras) and the
attenuation at higher frequencies which makes estimation particularly challenging.
CTF parameters, $\ctfparam$, are assumed to be given; CTF estimation
is beyond the scope of this work, but is routinely done using existing
tools, \eg, \citep{Mallick2005,Mindell2003}.

As noted above, and clearly seen in Figure \ref{fig:cryoimgs}, there is a
large amount of noise present in typical particle images.
This is primarily due to the sensitive nature of biological specimens,
requiring extremely low exposures.
The noise is modelled using an IID Gaussian distribution, resulting 
in the following expression for the conditional distribution 
of a particle image, $\img \in \R^{\N^2}$,
\begin{equation}
p(\img \,|\, \ctfparam, \projdir, \shiftdir, \density ) = \mathcal{N}(\img \,|\, \ctf{\ctfparam} \shift{\shiftdir} \proj{\projdir} \density, \noiseStd^2 \eye)
\label{eq:BaseLikelihood}
\end{equation}
where $\noiseStd$ is the standard deviation of the noise and
$\mathcal{N}(\cdot|\mu,\Sigma)$ is the multivariate normal distribution 
with mean vector $\mu$ and covariance matrix $\Sigma$.

In practice, due to computational considerations, 
Equation (\ref{eq:BaseLikelihood}) is evaluated in Fourier space, 
making use of the Fourier Slice Theorem and Parseval's Theorem to obtain
\begin{equation}
p(\ftimg \,|\, \ctfparam, \projdir, \shiftdir, \ftdensity ) = \mathcal{N}(\ftimg \,|\, \ftctf{\ctfparam} \ftshift{\shiftdir} \ftproj{\projdir} \ftdensity, \noiseStd^2 \eye)
\label{eq:BaseFTLikelihood}
\end{equation}
where $\ftimg$ is the 2D Fourier transform of the image, $\ftshift{\shiftdir}$
is the shift operator in Fourier space (a phase change), $\ftctf{\ctfparam}$ is
the CTF modulation in Fourier space (a diagonal operator), $\ftproj{\projdir}$ 
is a sinc interpolation operator which extracts a plane through the origin 
defined by the projection orientation $\projdir$ and $\ftdensity$ is the
3D Fourier transform of $\density$.
To speed the computation of the likelihood, and due to the level of
noise and attenuation of high frequencies by the CTF, a maximum frequency
is specified, $\rad$, beyond which frequencies are ignored.

The 3D pose, $\projdir$, and shift, $\shiftdir$, of each particle image
are unknown and treated as latent variables which are marginalized out 
\citep{Sigworth1998,Scheres2007}.
Assuming $\projdir$ and $\shiftdir$ are independent of each other and
the density $\density$, one obtains
\begin{equation}
p(\ftimg \,|\, \ctfparam, \ftdensity) = \int_{\R^2} \int_{\SO} p(\ftimg | \ctfparam, \projdir, \shiftdir, \ftdensity ) p(\projdir) p(\shiftdir) d\projdir d\shiftdir
\label{eq:Marginalization}
\end{equation}
where $p(\projdir)$ is a prior over 3D poses, $\projdir \in \SO$, and 
$p(\shiftdir)$ is a prior over translations, $\shiftdir \in \R^2$.
In general, nothing is known about the projection direction so $p(\projdir)$ 
is assumed to be a uniform distribution.
Particles are picked to be close to the center of each image, so 
$p(\shiftdir)$ is chosen to be a Gaussian distribution centered in the image.
The above double integral is not analytically tractable, so numerical
quadrature is used \citep{Lebedev1999,Graef2009}.
The conditional probability of an image (likelihood) then becomes
\begin{equation}
p(\ftimg | \ctfparam, \ftdensity) \approx \sum_{j=1}^{M_\projdir} w_j^{\projdir} \sum_{\ell=1}^{M_\shiftdir} w_k^{\shiftdir} p(\ftimg | \ctfparam, \projdir_j, \shiftdir_\ell, \ftdensity ) p(\projdir) p(\shiftdir)
\label{eq:ApproxMarginalization}
\end{equation}
where $\{(\projdir_j,w_j^{\projdir})\}_{j=1}^{M_\projdir}$ are weighted
quadrature points over $\SO$ and
$\{(\shiftdir_\ell,w_\ell^{\shiftdir})\}_{\ell=1}^{M_\shiftdir}$ are 
weighted quadrature points over $\R^2$.
The accuracy of the quadrature scheme, and consequently the values of
$M_\projdir$ and $M_\shiftdir$,  are set automatically based on $\rad$,
the specified maximum frequency such that higher values of $\rad$ results
in more quadrature points.

Given a set of $K$ images with CTF parameters
$\data = \{(\img_i,\ctfparam_i)\}_{i=1}^{K}$ and assuming conditional
independence of the images, the posterior probability of a density $\density$ is
\begin{equation}
p(\density | \data) \propto p(\density) \prod_{i=1}^K p(\ftimg_i | \ctfparam_i, \ftdensity)
\label{eq:Posterior}
\end{equation}
where $p(\density)$ is a prior over 3D molecular electron densities.
Several choices of prior are explored below, but we found that
a simple independent exponential prior worked well.
Specifically, $p(\density) = \prod_{i=1}^{\N^3} \lambda e^{-\lambda\density_i}$
where $\density_i$ is the value of the $i$th voxel and $\lambda$ is the 
inverse scale parameter.
Other choices of prior are possible and is a promising direction for
future research.

Estimating the density now corresponds to finding $\density$
which maximizes Equation (\ref{eq:Posterior}).
Taking the negative log and dropping constant factors, the optimization 
problem becomes $\arg\min_{\density \in \R_+^{\N^3}} \obj(\density)$,
\begin{equation}
\obj(\density) = -\log p(\density) - \sum_{i=1}^K\log p(\ftimg_i | \ctfparam_i, \ftdensity)
\label{eq:OptimProblem}
\end{equation}
where $\density$ is restricted to be positive (negative density is 
physically unrealistic).
Optimizing Eq.\ (\ref{eq:OptimProblem}) directly is costly due to the 
marginalization in Eq.\ (\ref{eq:ApproxMarginalization}) as well as the
large number ($K$) of particle images in a typical dataset.
To deal with these challenges, the following sections propose the use
of two techniques, namely, stochastic optimization and importance sampling.

\subsection{Stochastic Optimization}
In order to efficiently cope with the large number of particle images in a
typical dataset, we propose the use of stochastic optimization methods.
Stochastic optimization methods exploit the large amount of redundancy
in most datasets by only considering subsets of data (\ie, images)
at each iteration by rewriting the objective as
$\obj(\density)=\sum_k \obj_k(\density)$ where each $\obj_k(\density)$
evaluates a subset of data.
This allows for fast progress to be made before a batch optimization algorithm
would be able to take a single step.

There are a wide range of such methods, ranging from simple stochastic 
gradient descent with momentum \citep{Nesterov1983,Polyak1964,Sutskever2013}
to more complex methods such as Natural Gradient methods
\citep{Amari2000,Amari1998,LeRoux2010,LeRoux2008}
and Hessian-free optimization \citep{Martens2010}.
Here we propose the use of Stochastic Average Gradient Descent (SAGD) 
\citep{LeRoux2012} which has several important advantages.
First, it is effectively self-tuning, using a line-search to determine and
adapt the learning rate.
This is particularly important, as many methods require significant 
manual tuning for new objective functions and, potentially, each new
dataset.
Further, it is specifically designed for the finite dataset case allowing
for faster convergence.

At each iteration $\iter$, SAGD \citep{LeRoux2012} considers
only a single subset of data, $k_{\iter}$, which defines
part of the objective function
$\obj_{k_{\iter}}(\density)$ and its gradient $\dobj_{k_{\iter}}(\density)$. 
The density $\density$ is then updated as
\begin{equation}
\density_{\iter+1} = \density_{\iter} - \frac{\epsilon}{K L} \sum_{j=1}^K \densitygrad_j^\iter
\end{equation}
where $\epsilon$ is a base learning rate, $L$ is a Lipschitz constant
of $\dobj_k(\density)$, and
\begin{equation}
\densitygrad_k^\iter =
\begin{cases}
\dobj_k(\density_{\iter}) & k = k_{\iter} \\
\densitygrad_k^{\iter-1} & \mbox{otherwise}
\end{cases} 
\label{eq:sagd_grad}
\end{equation}
is the most recent gradient evaluation of datapoint $j$ at iteration $\iter$.
This step can be computed efficiently by storing the gradient of each
observation and updating a running sum each time a new gradient is seen.
The Lipschitz constant $L$ is not generally known but can be estimated
using a line-search technique.
Theoretically, convergence occurs for values of 
$\epsilon \leq \frac{1}{16}$ \citep{LeRoux2012}, however in practice larger
values at early iterations can be beneficial, thus we use
$\epsilon = \max(\frac{1}{16}, 2^{1-\lfloor \iter/150 \rfloor})$.
To allow parallelization and reduce the memory requires of SAGD,
the data is divided into minibatches of 200 particles images.
Finally, to enforce the positivity of density, negative values of
$\density$ are truncated to zero after each iteration.
More details of the stochastic optimization can be found in the Supplemental Material.

\subsection{Importance Sampling}
While stochastic optimization allows us to scale to large datasets, the cost of computing
the required gradient for each image remains high due to the marginalization over 
orientations and shifts.
Intuitively, one could consider randomly selecting a subset of the terms in 
Eq.\ (\ref{eq:ApproxMarginalization}) and using this as an approximation.
This idea is formalized by importance sampling (IS) which allows
for an efficient and accurate approximation of the discrete sums in
Eq.\ (\ref{eq:ApproxMarginalization}).\footnote{One can also apply
importance sampling directly to the continuous integrals in
Eq.\ (\ref{eq:Marginalization}) but it can be computationally advantageous
to precompute a fixed set of projection and shift matrices, $\ftproj{\projdir}$
and $\ftshift{\shiftdir}$, which can be reused across particle images.}
A full review of importance sampling is beyond the scope of this paper
but we refer readers to \citep{Tokdar2010}.

To apply importance sampling, consider the inner sum from 
Eq.\ (\ref{eq:ApproxMarginalization}), rewriting it as
\begin{equation}
\phi_j^{\projdir} = \sum_{\ell=1}^{M_{\shiftdir}} w_\ell^{\shiftdir} p_{j,\ell} 
= \sum_{\ell=1}^{M_{\shiftdir}} \isp_\ell^{\shiftdir} \left( \frac{w_\ell^{\shiftdir} p_{j,\ell}}{\isp_\ell^{\shiftdir}} \right)
\end{equation}
where $p_{j,\ell} =  p(\ftimg | \ctfparam, \projdir_j, \shiftdir_\ell, \ftdensity) p(\projdir_j) p(\shiftdir_\ell)$
and $\isP^{\shiftdir} = (\isp_1^{\shiftdir}, \dots, \isp_{M_\shiftdir}^{\shiftdir})^T$
is the parameter vector of a multinomial importance distribution such that
$\sum_{\ell=1}^{M_\shiftdir} \isp_\ell^{\shiftdir} = 1$ and $\isp_\ell^{\shiftdir}>0$.
The domain of $\isP^{\shiftdir}$ corresponds to the set of quadrature
points in Equation (\ref{eq:ApproxMarginalization}).
Then, $\phi_j^{\projdir}$ can be thought of as the expected value 
$E_\ell[ \frac{w_\ell^{\shiftdir} p_{j,\ell}}{\isp_\ell^{\shiftdir}}]$
where $\ell$ is a random variable distributed according to $\isP^{\shiftdir}$.
If a set of $N_{\shiftdir} \ll M_{\shiftdir}$ random indexes $\mathfrak{I}^{\shiftdir}$ are 
drawn according to $\isP^{\shiftdir}$, then
\begin{equation}
\phi_j^{\projdir} \approx \frac{1}{N_{\shiftdir}} \sum_{\ell \in \mathfrak{I}^{\shiftdir}} \frac{w_\ell^{\shiftdir} p_{j,\ell}}{\isp_\ell^{\shiftdir}}\ .
\end{equation}
Thus, we can efficiently approximate $\phi_j^{\projdir}$ by drawing samples 
according to the importance distribution $\isP^{\projdir}$ and
computing the average.
Using this approximation in Eq.\ (\ref{eq:ApproxMarginalization}) gives
\begin{equation}
p(\ftimg | \ctfparam, \ftdensity) \approx \sum_{j=1}^{M_\projdir} w_j^{\projdir} \frac{1}{N_{\shiftdir}} \left( \sum_{\ell \in \mathfrak{I}^{\shiftdir}} \frac{w_\ell^{\shiftdir} p_{j,\ell}}{\isp_\ell^{\shiftdir}} \right)
\end{equation}
and importance sampling can be similarly used for the outer summation to give
\begin{equation}
p(\ftimg | \ctfparam, \ftdensity) \approx 
\sum_{j \in \mathfrak{I}^{\projdir}}
\frac{w_j^{\projdir}}{N_{\projdir} \isp_j^{\projdir}}
\left( \sum_{\ell \in \mathfrak{I}^{\shiftdir}}
\frac{w_\ell^{\shiftdir}}{N_{\shiftdir} \isp_\ell^{\shiftdir}}
p_{j,\ell} \right)
\label{eq:ISApprox}
\end{equation}
where $\mathfrak{I}^{\projdir}$ are samples drawn from the importance
distribution $\isP^{\projdir} = (\isp_1^{\projdir}, \dots, \isp_{M_\projdir}^{\projdir})^T$
used for approximating 
\begin{equation}
\phi_\ell^{\shiftdir}
= \sum_{j=1}^{M_{\projdir}} w_j^{\projdir} p_{j,\ell} 
\approx \frac{1}{N_{\projdir}} \sum_{j \in \mathfrak{I}^{\projdir}} \frac{w_j^{\projdir} p_{j,\ell}}{\isp_j^{\projdir}} \ .
\end{equation}
The accuracy of the approximation in Eq.\ (\ref{eq:ISApprox}) is controlled
by the number of samples used, with the error going to zero as $N$ 
increases.
We use $N = s_0 s(\isP)$ samples where 
$s(\isP) = \left( \sum_\ell \isp_\ell^2 \right)^{-1}$  is the effective
sample size \citep{Doucet2000} and $s_0$ is a scaling factor.
This choice ensures that when the importance distribution is diffuse,
more samples are used.

\begin{figure}
\centering
\includegraphics[width=0.3\textwidth,clip,trim=0 0 0 0]{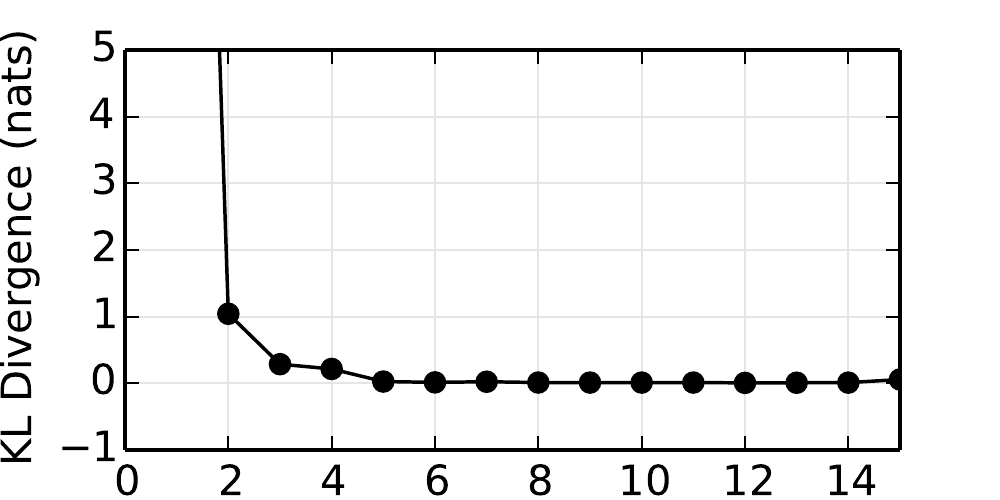}
\caption{\label{fig:kldiv}
The KL divergence between the 
values of $\phi^{\projdir}$ at the current
and previous epochs on the thermus dataset.
}
\end{figure}

While the estimates provided by IS are unbiased, their error can be 
arbitrarily bad if the importance distribution is not well chosen.
To choose a suitable importance distribution, we make two observations.
First, the values $\phi_\ell^\shiftdir$ and $\phi_j^\projdir$ are proportional
to the marginal probability of single particle image having been generated
with shift $\shiftdir_\ell$ or pose $\projdir_j$, making them natural choices
on which to base the importance distributions.
Second, these values remain stable once the rough shape of the structure has 
been determined.
This can be seen in Figure \ref{fig:kldiv} which shows that by the
third epoch the KL divergence of the values of $\phi^{\projdir}$ from one epoch to
the next is extremely small.

\setlength{\tabcolsep}{1.5pt}

\begin{figure}
\centering
\begin{tabular}{ccc}
\includegraphics[height=0.2\textwidth,clip,trim=220 120 220 120]{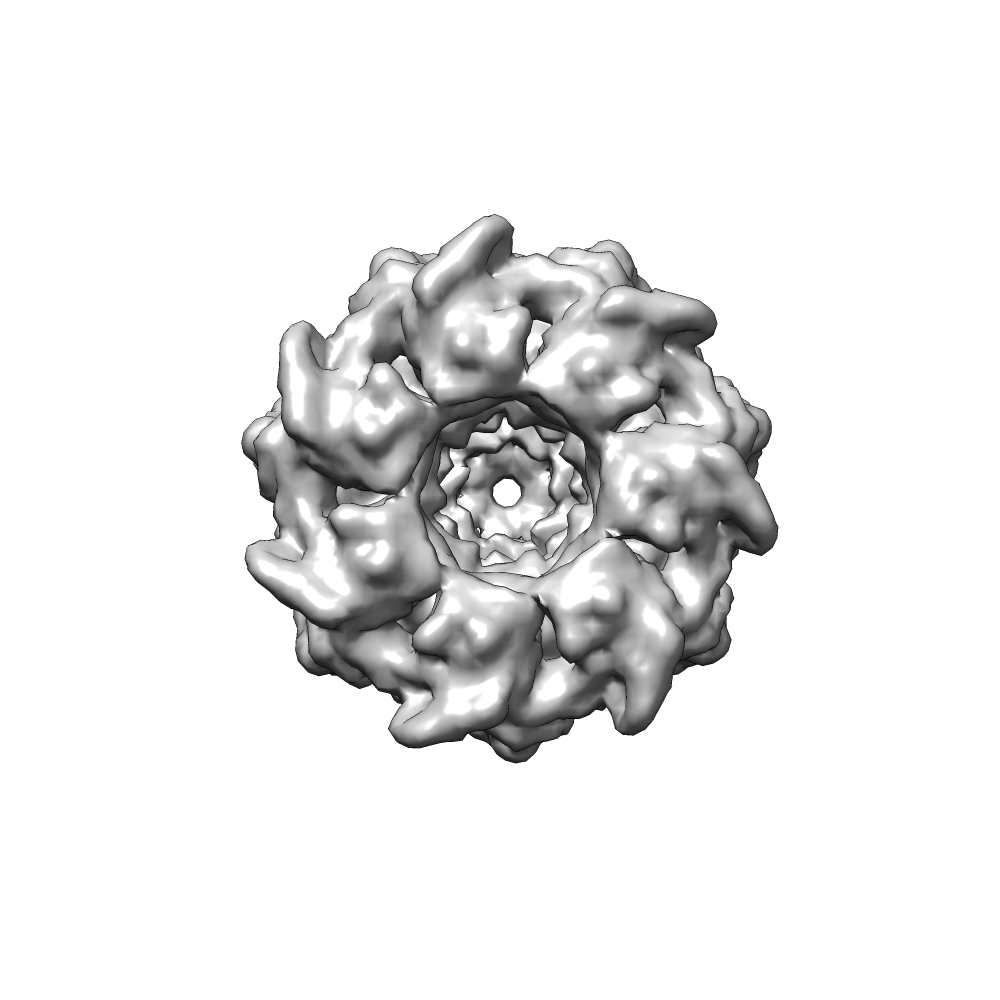} &
\includegraphics[height=0.2\textwidth,clip,trim=260 120 260 120]{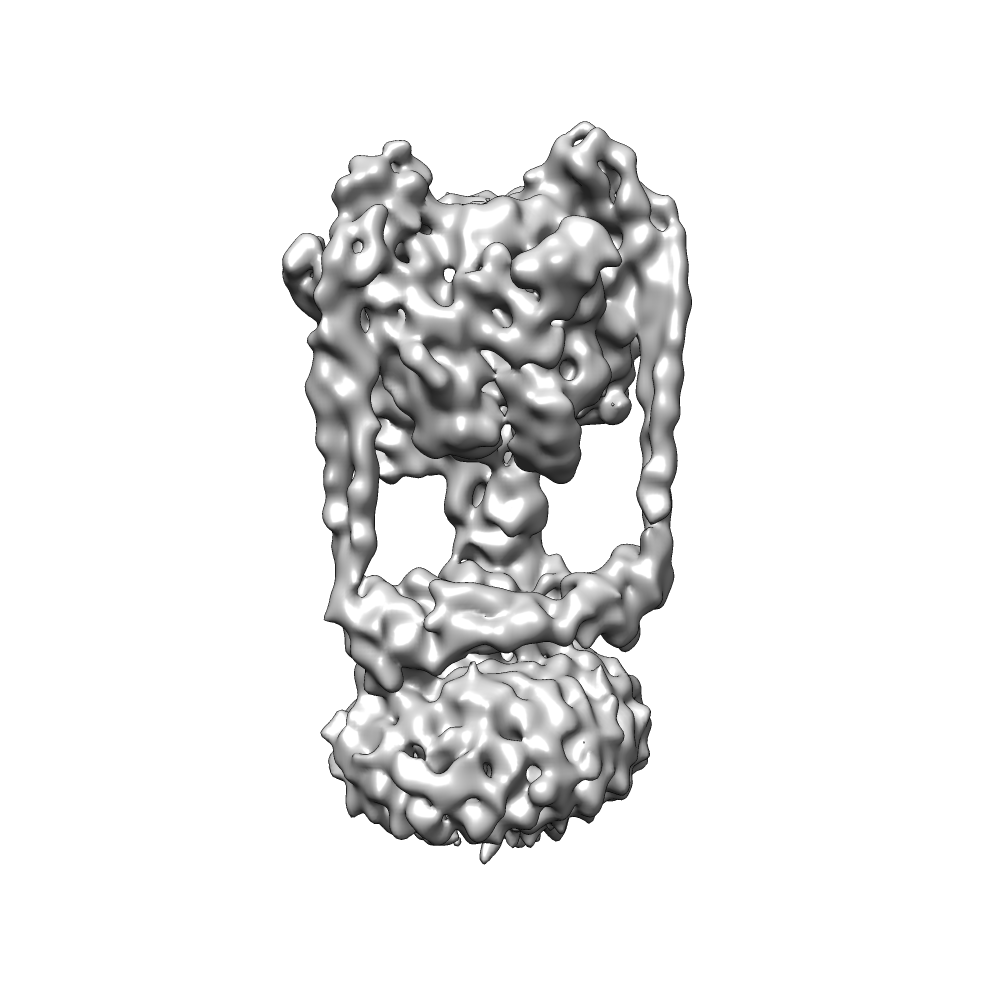} &
\includegraphics[height=0.2\textwidth,clip,trim=260 120 260 120]{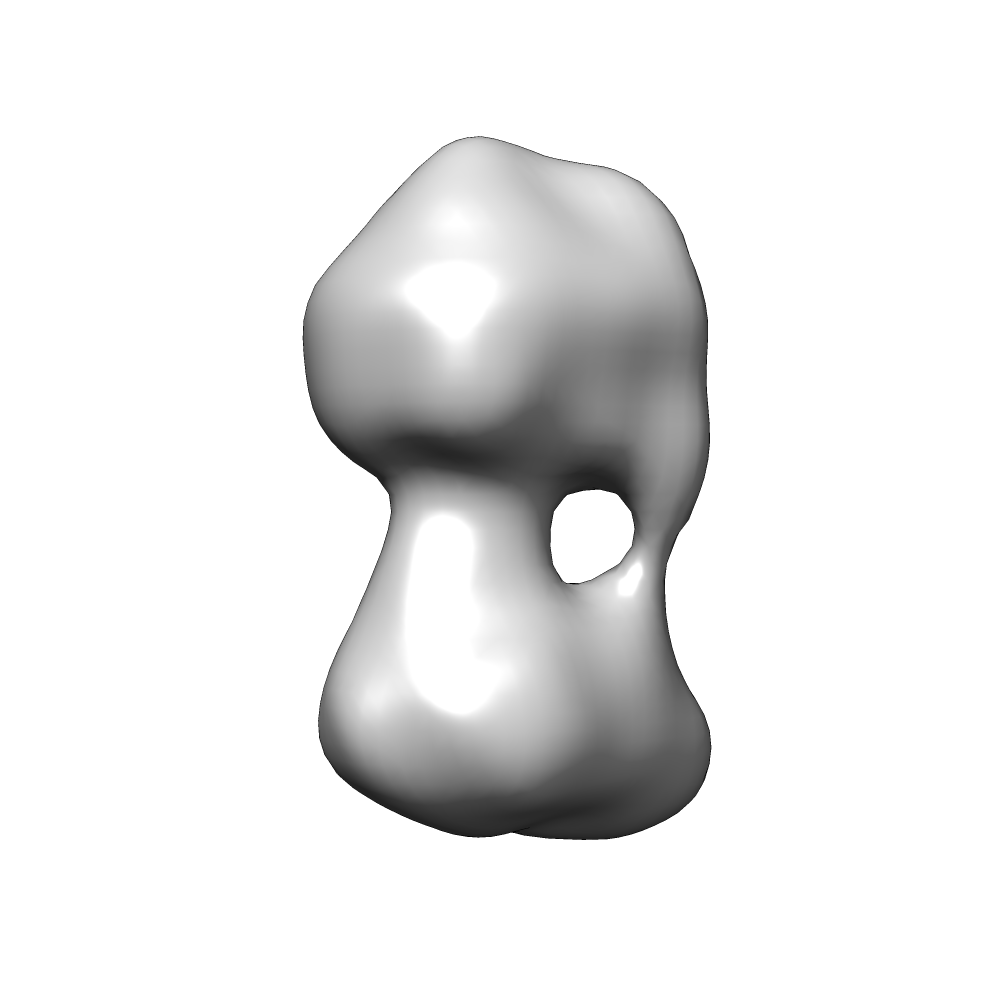} \\
{\small GroEL-GroES \citep{Xu1997}} &
{\small Thermus ATPase \citep{Lau2012}} &
{\small Bovine ATPase \citep{Rubinstein2003}}
\end{tabular}
\caption{\label{fig:groundtruths}
Previously published structures for the datasets used in this paper.
}
\end{figure}

Thus we use these quantities, computed from the previous iterations, to construct 
the importance distributions at the current iteration.
Dropping the $\projdir$ or $\shiftdir$ superscripts for clarity, let
$\mathfrak{I}$ be the set of samples evaluated at the previous iteration and
$\phi_{i}$ be the computed values for $i\in\mathfrak{I}$.
Then the importance distribution used at the current iteration is
\begin{eqnarray}
\isp_j = (1-\alpha) Z^{-1} \hat{\phi}_j + \alpha \psi_j
\label{eq:impsamp_mixing}
\end{eqnarray}
where $\psi_j$ is a uniform prior distribution, $\alpha$ controls how much
the previous distribution is relied on, $Z = \sum_j \hat{\phi}_j$, 
and $\hat{\phi}_j = \sum_{i\in\mathfrak{I}} \phi_{i}^{1/T} \kernmat_{i,j}$.
Here $T$ is an annealing parameter and $\kernmat_{i,j}$ are entries of a kernel matrix
computed on the quadrature points which diffuses probability to nearby quadrature points.
The values for $\alpha = \max(0.05,2^{-0.25\lfloor {\iter}_{prev}/50 \rfloor})$ and
$T = \max(1.25, 2^{10.0/\lfloor {\iter}_{prev}/50 \rfloor})$ are set so that
at early iterations, when the underlying density is changing, we rely more heavily
on the prior.
For $\kernmat$ we use a Gaussian kernel for the shifts and a Fisher
kernel for the orientations.
The bandwidth of the kernel is tuned based on the current resolution of the
quadrature scheme, \eg, the Gaussian shift kernel bandwidth is set to be equal
to the spacing between the shift quadrature points.
More details on importance sampling can be found in the Supplemental Material.

\section{Experiments}
\begin{figure}
\centering
\begin{tabular}{ccccc}
\includegraphics[width=0.09\textwidth,clip,trim=0 -250 0 0]{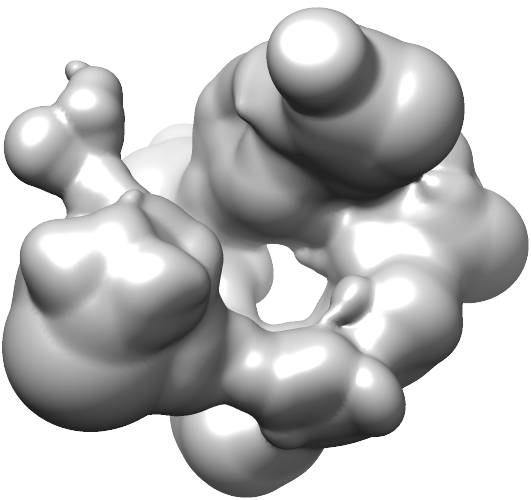} &
\includegraphics[width=0.09\textwidth]{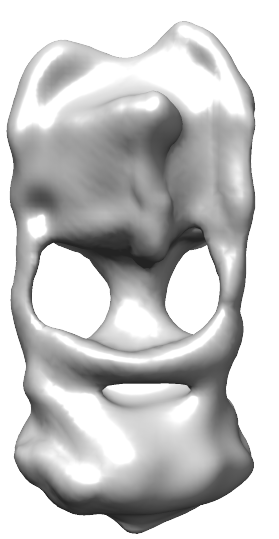} &
\includegraphics[width=0.09\textwidth]{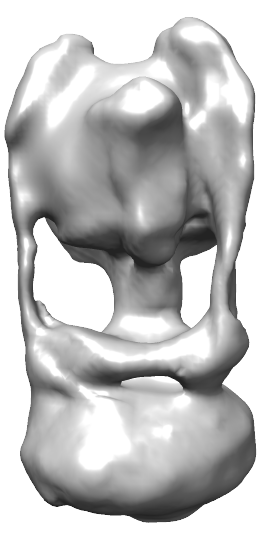} &
\includegraphics[width=0.09\textwidth]{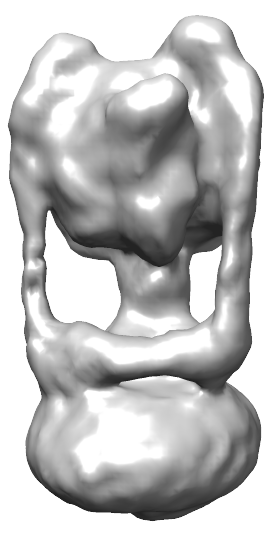} &
\includegraphics[width=0.09\textwidth]{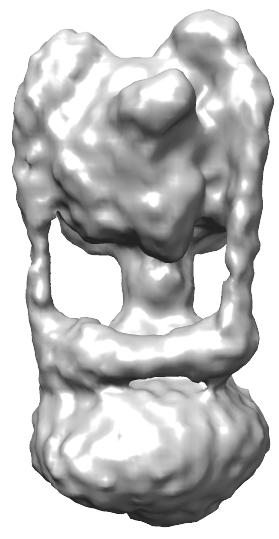} \\ 
{\small Initial} &
{\small 1 hour} &
{\small 6 hours} &
{\small 12 hours} &
{\small 24 hours}
\end{tabular}
\caption{\label{fig:iterations}
The random initialization (left) used in all experiments,
generated by summing random spheres, and 
reconstruction of the \emph{thermus} dataset after various
amounts of computation.
Note that within an hour of computation, the gross structure
is already well determined, after which fine details emerge 
gradually.
}
\end{figure}

The proposed method was applied to two experimental datasets and
one synthetic dataset.
All experiments used the same parameters and were initialized
using the same randomly generated density shown in
Figure \ref{fig:iterations}(left).
The maximum frequency considered was gradually increased 
from an initial value of $\rad = 1/40$\angs\ to 
a maximum of $\rad = 1/10$\angs.
This maximum frequency corresponds to the resolution of the 
best published results for the datasets used here,
\ie, \citep{Lau2012}.
Optimizations were run until the maximum resolution was reached and the average
error on a held-out set of $100$ particle images stopped improving,
around 5000 iterations.

\setlength{\tabcolsep}{2pt}

\paragraph{Datasets}
The first dataset was ATP synthase from the \emph{thermus thermophilus}
bacteria, a large transmembrane molecule.
The \emph{thermus} dataset consisted of $46,105$ particle images which
were provided by \citet{Lau2012}.
The high resolution structure from \citep{Lau2012} and some sample images 
are shown in Figure \ref{fig:cryoimgs}.
The second dataset was \emph{bovine} mitochondrial ATP synthase
\citep{Rubinstein2003}.
The \emph{bovine} dataset, provided by \citet{Rubinstein2003}, consisted of
$5,984$ particle images.
In all cases the particle images provided were $128 \times 128$,
had a resolution of $2.8$\angs\ ($0.28nm$ per pixel) and CTF
information for each particle image was provided.
The noise level, $\sigma$, was estimated by computing the standard deviation
of pixels around the boundary of the particle images.

To showcase the ability of our method to handle a dramatically different type
of structure, a third dataset was synthesized by taking an existing structure
from the Protein Data Bank\footnote{Structure 1AON from \url{http://pdb.org}},
GroEL-GroES-(ADP)7 \citep{Xu1997}, and generating $40,000$ random projections
according to the generative model.
CTF, signal-to-noise level and other parameters were set realistically based on 
the \emph{thermus} dataset values.
This structure, as well as previously solved structures of the bovine and 
thermus ATP synthase molecules are shown in Figure \ref{fig:groundtruths}.
GroEL-GroES, was selected because it is
structurally unlike either of the bovine or thermus ATP synthase
molecules.
Sample GroEL synthetic images can be see in Figure \ref{fig:resgrid} (top 
left).

Results of our method on these datasets are shown in Figure \ref{fig:resgrid}.
Sample particle images are shown, along with an iso-surface and slices of
the final estimated density.
Computing these reconstructions took less than 24 hours in all cases.
Further, even at early iterations, reasonable structures are available.
Figure \ref{fig:iterations} shows the estimated structure for the
\emph{thermus} dataset over time during optimization.
Notably, after just one hour (during which only a fraction of the full
dataset is seen), the low-resolution shape of the structure has already
been determined.

\begin{figure}
\centering
\includegraphics[width=0.47\textwidth]{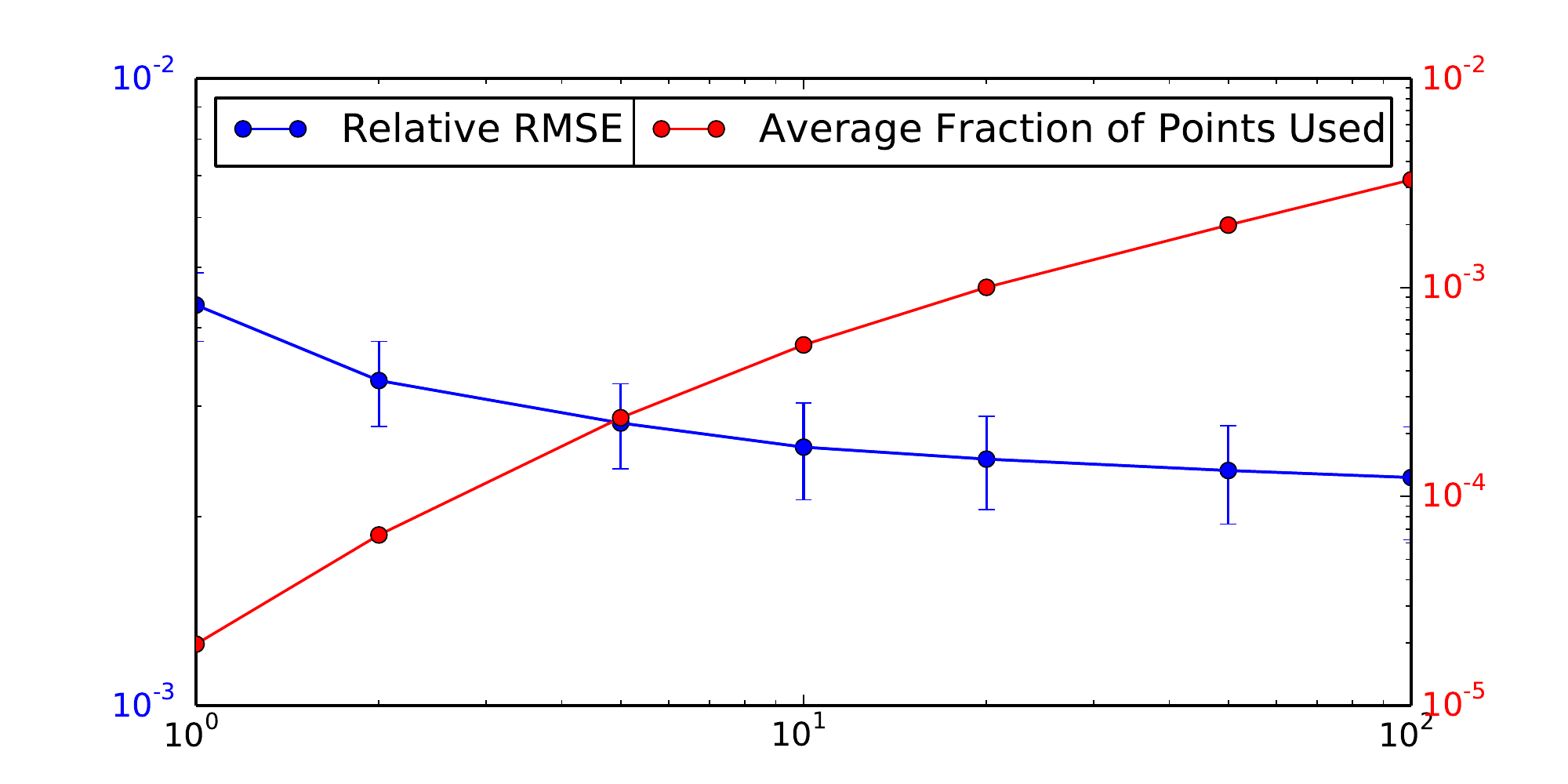}
\caption{\label{fig:esserrs}
Relative error (blue, left axis) and fraction of total quadrature points (red, right axis) used
in computing $\log p(\ftimg | \ctfparam, \ftdensity)$
as a function of the ESS scaling factor, $s_0$ (horizontal axis).
Note the log-scale of the axes.
}
\end{figure}

\paragraph{Importance Sampling}
To validate our importance sampling approach we evaluated the error made in
computing $\log p(\ftimg | \ctfparam, \ftdensity)$ using IS against computing 
the exact sum in Equation (\ref{eq:ApproxMarginalization}) without IS.
This error is plotted in Figure \ref{fig:esserrs}, along with the fraction of
quadrature points used at various values of $s_0$.
Based on these plots we selected a factor of $s_0=10$ for all experiments as 
a trade-off between accuracy and speed achieving a relative error of less then 
$0.1\%$ while still providing significant speedups.

\begin{figure}
\centering
\includegraphics[width=0.5\textwidth]{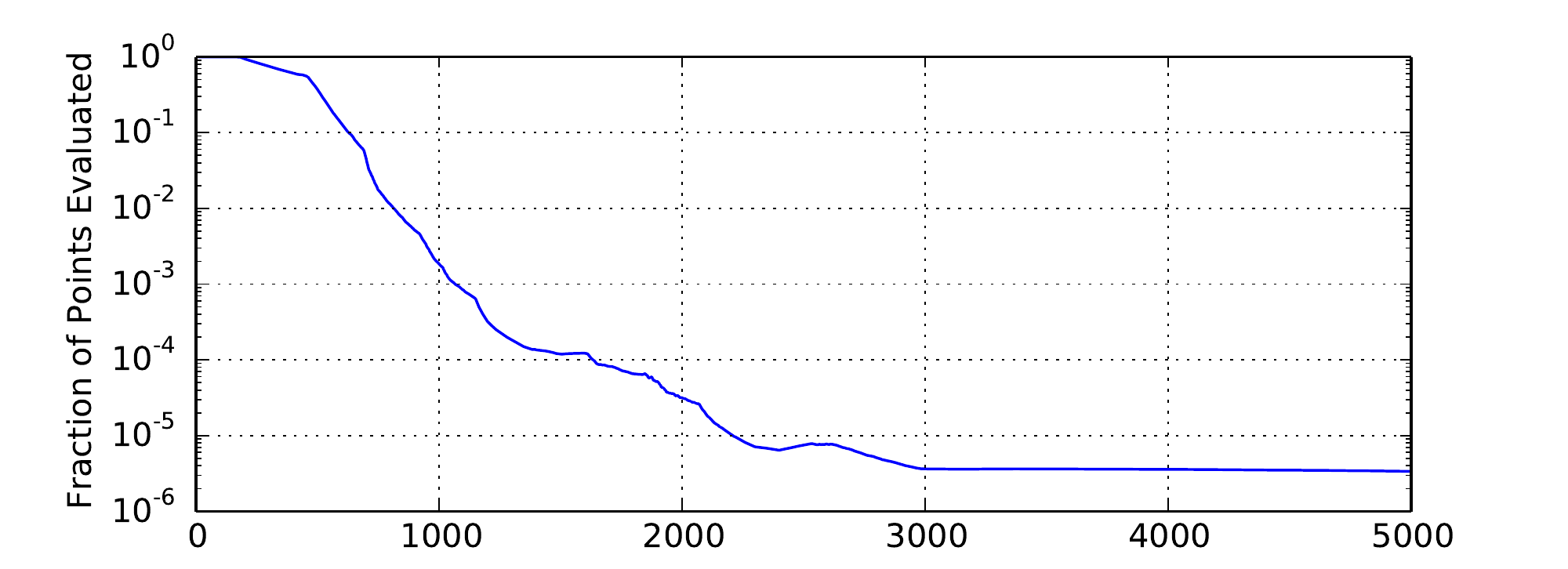}
\caption{\label{fig:speedups}
The fraction of quadrature points evaluated on average
during optimization (horizontal axis is iterations).
As resolution increases, the speedup obtained increases 
significantly yielding more than a 100,000 fold speedup.
}
\end{figure}

To see just how much of a speedup importance sampling provides in practice, we
plotted in Figure \ref{fig:speedups} the fraction of quadrature points which
needed to be evaluated during optimization.
As can be seen initially, all quadrature points are evaluated but as 
optimization progresses and the density (and consequently the distribution
over poses) becomes better determined  importance sampling yields larger and
larger speedups.
At the full resolution, importance sampling provided more than a 100,000 fold
speedup.

\begin{figure}
\centering
\includegraphics[height=0.18\textwidth,width=0.42\textwidth,clip,trim=74 31 148 32]{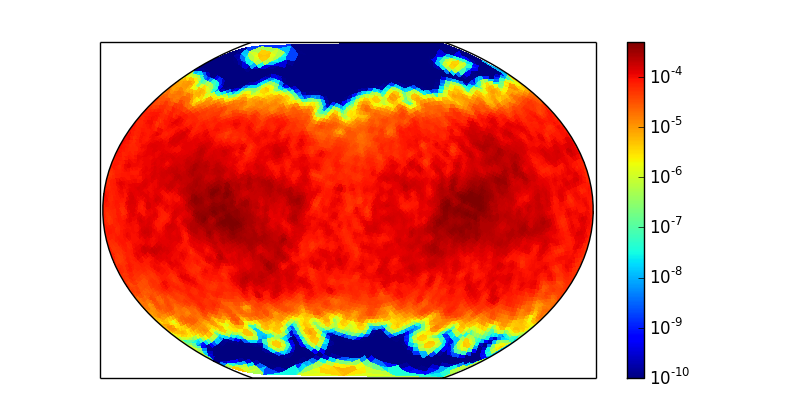}%
\includegraphics[height=0.18\textwidth,clip,trim=434 20 20 32]{figs/thermus_avgphirs.png}
\caption{\label{fig:avgisdist}
A Winkel-Tripel projection of the importance distribution of view directions, $\isP^{\projdir}$,
averaged over the \emph{thermus} dataset at a typical iteration.
Clearly visible is the equatorial belt of likely views, while
axis aligned views (those on the top or bottom of the plot)
are rarely seen.
\vspace*{-0.25cm}
}
\end{figure}

No prior knowledge of the orientation distribution was assumed.
However, for many particles, certain views are more likely than others.
This fact can be seen by examining the average importance distribution 
for the thermus dataset, shown in Figure \ref{fig:avgisdist} for a typical
iteration.
Here we can see clearly that the distribution of views forms an equatorial
belt around the particle, while top or bottom views are rarely if ever seen.
This phenomenon is well known for particles like these (\eg, see
\citep{Rubinstein2003} where this knowledge was used directly in estimation),
validating our sampling approach and suggesting a use of this average 
importance distribution to supplement the uniform prior distribution in 
Eq.\ (\ref{eq:impsamp_mixing}).

\begin{figure}
\centering
\begin{tabular}{ccc}
\includegraphics[height=0.14\textwidth]{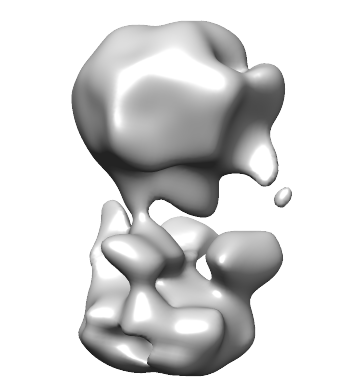} &
\includegraphics[height=0.14\textwidth]{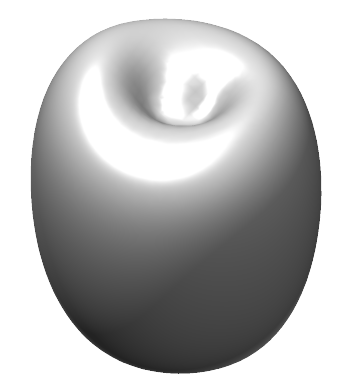} &
\includegraphics[height=0.14\textwidth,clip,trim=120 120 120 120]{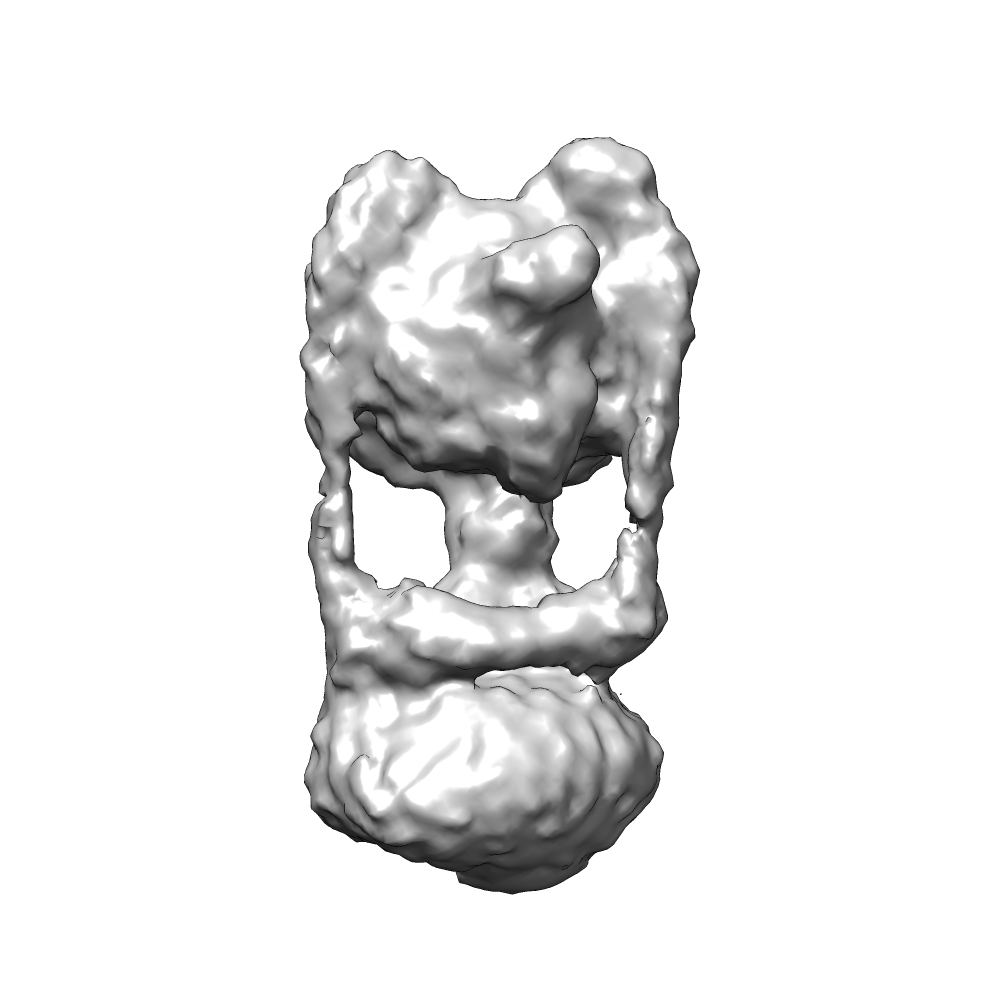} \\
{\small Projection Matching} &
{\small RELION} &
{\small Proposed Approach}
\end{tabular}
\caption{\label{fig:baselines}
Baseline comparisons to two existing standard methods.
Iterative projection matching and reconstruction (left) and
RELION \citep{Scheres2012} (middle).
The proposed method (right) is able to determine the correct structure
while projection matching and RELION both become trapped in poor local 
optima.
See Fig.\ \ref{fig:baselines}(middle) for comparison.
All methods used the same random initialization shown in Fig.\ \ref{fig:iterations}.
}
\end{figure}

\paragraph{Initialization and Comparison to State-of-the-Art}
To compare this method to existing methods for structure determination, 
we selected two representative approaches.
The first is a standard iterative projection matching scheme where
images are matched to an initial density through a global cross-correlation
search \citep{Grigorieff2007}.
The density is then reconstructed based on these fixed orientations and this
process is iterated.
The second is the RELION package described in \citep{Scheres2012} which
uses a similar marginalized model as our method but with a batch
EM algorithm to perform optimization.
We used publicly available code for both of these approaches on the
\emph{thermus} dataset and initialized using the density shown in Figure
\ref{fig:iterations}.
We ran each method for a number of iterations roughly equivalent
computationally to the 5000 iterations used by our method and the results
are shown in Figure \ref{fig:baselines}.
In both cases the approaches had clearly determined an incorrect structure
and appeared to have converged to a local minimum as no further progress was
made.
Both projection matching and RELION have been used successfully for
reconstruction by others and are not recommended to be used without a good
initialization.
Our results support this recommendation as neither approach converges from
random initializations.
In practice, it is difficult to construct good initializations for molecules
of unknown shape \cite{Henderson2012}, giving our proposed method a
significant advantage.

\begin{figure}
\centering
\includegraphics[width=0.15\textwidth,clip,trim=50 30 125 0]{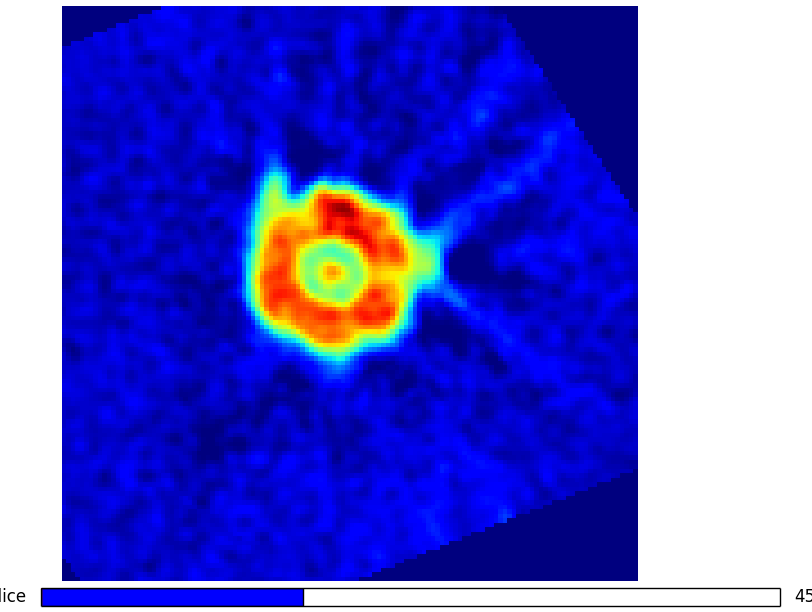} \hfill
\includegraphics[width=0.15\textwidth,clip,trim=50 30 125 0]{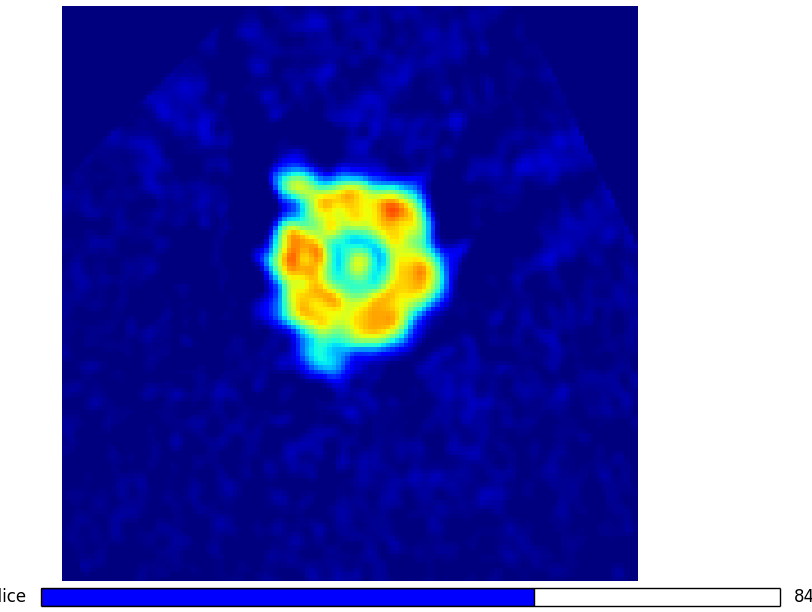} \hfill
\includegraphics[width=0.15\textwidth,clip,trim=50 30 125 0]{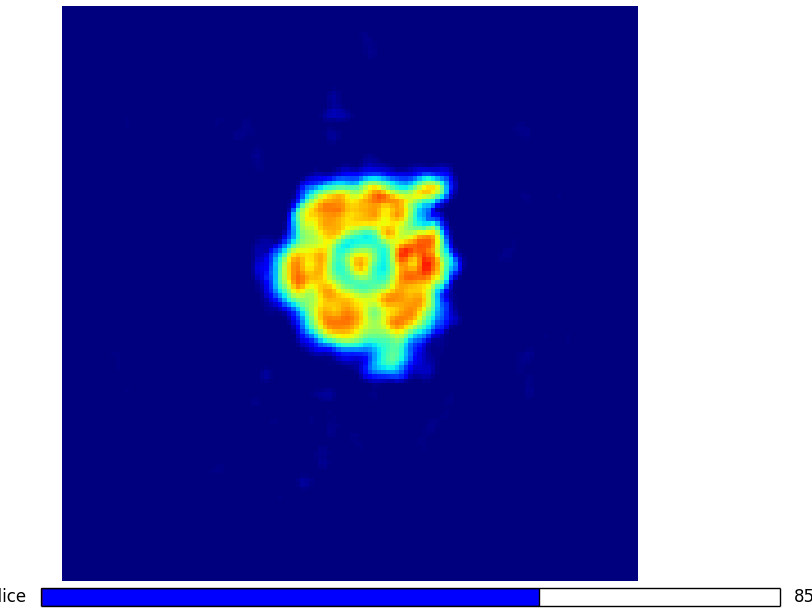}
\caption{\label{fig:priors}
Slices through the reconstructions with (from left to right)
uniform, CAR and exponential priors.
The exponential prior does the best job of suppressing noise
in the background without oversmoothing fine details within
the structure.
Blue corresponds to small or zero density and red corresponds
to high density.
\vspace*{-0.25cm}
}
\end{figure}

\setlength{\tabcolsep}{2pt}
\newcommand{\dataname}[1]{\rotatebox{90}{\small \bf #1}}

\begin{figure*}
	\centering
	\begin{tabular}{c|c|c|c}
	&
	\bf{Sample Particle Images} &
	\bf{3D Reconstruction} &
	\bf{3D Slices} \\
	\hline
	\dataname{\ \ \ \ \ \ \ GroEL-GroES} &
	\begin{subfigure}[b]{0.29\textwidth}
	\includegraphics[width=0.32\textwidth]{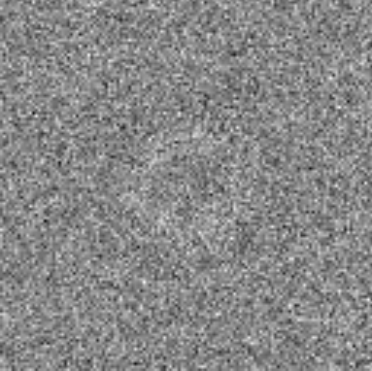}
	\includegraphics[width=0.32\textwidth]{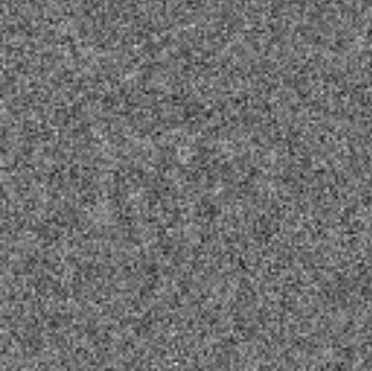}
	\includegraphics[width=0.32\textwidth]{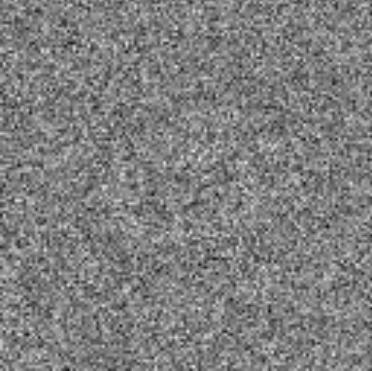}\\
	\includegraphics[width=0.32\textwidth]{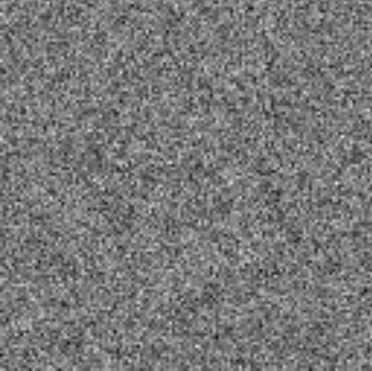}
	\includegraphics[width=0.32\textwidth]{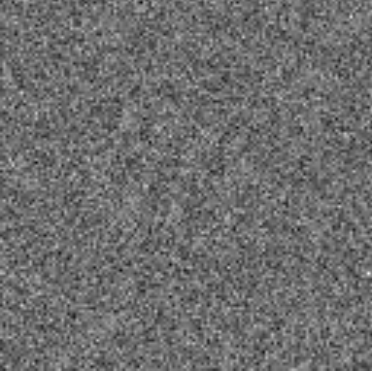}
	\includegraphics[width=0.32\textwidth]{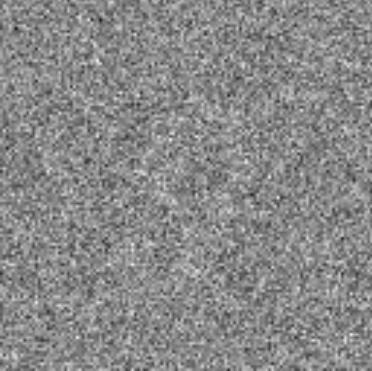}
	\end{subfigure}%
	&
	\includegraphics[width=0.17\textwidth,clip,trim=175 120 175 120]{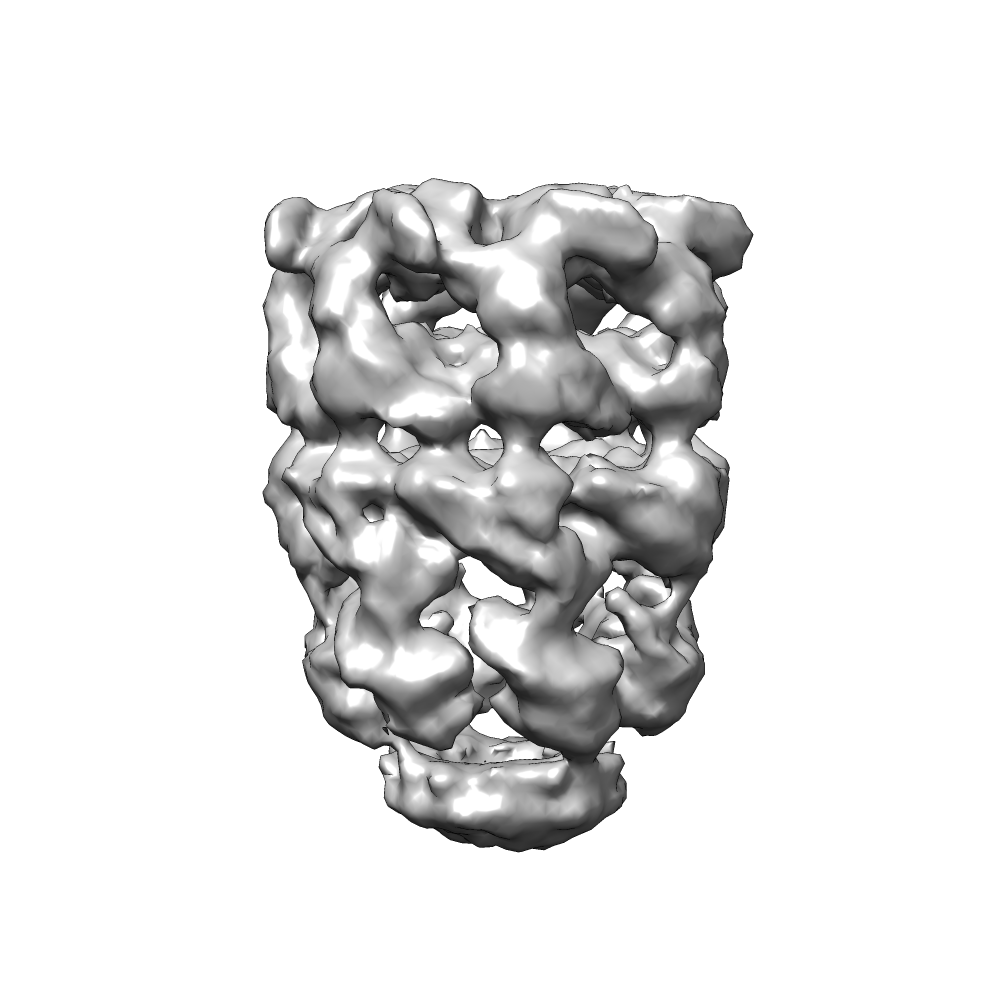}%
	\includegraphics[width=0.17\textwidth,clip,trim=175 120 175 120]{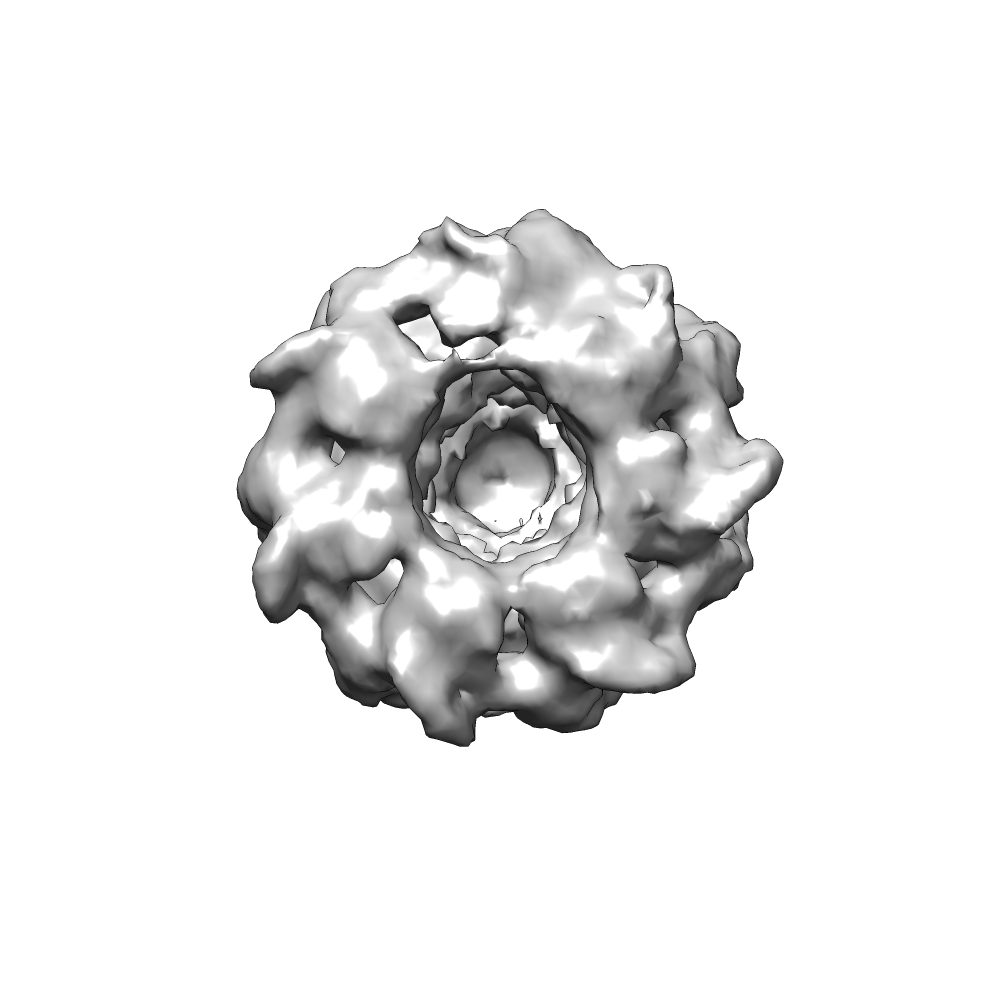}%
	&
	\includegraphics[width=0.17\textwidth,clip,trim=150 120 150 120]{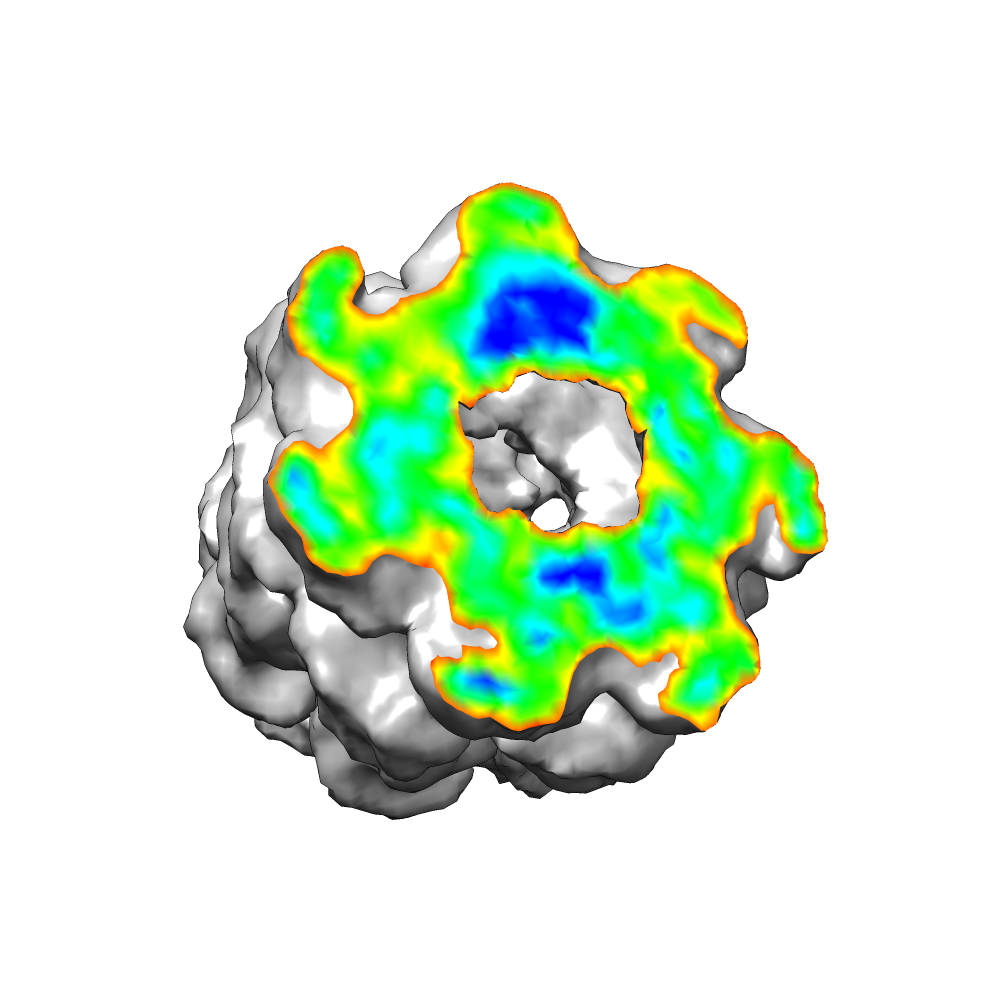}%
	\includegraphics[width=0.17\textwidth,clip,trim=150 120 150 120]{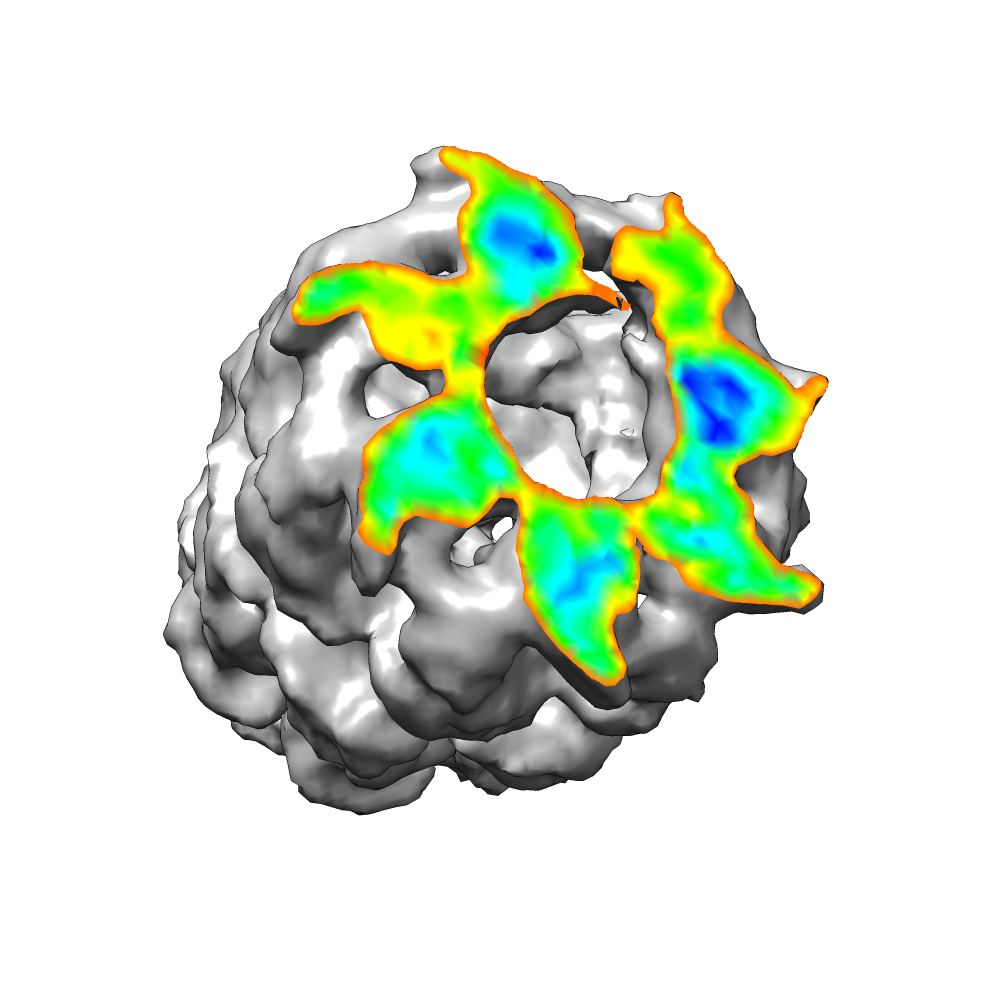}\\%
	\hline
	\dataname{\ \ \ \ \ \ Thermus ATPase} &
	\begin{subfigure}[b]{0.29\textwidth}
	\includegraphics[width=0.32\textwidth]{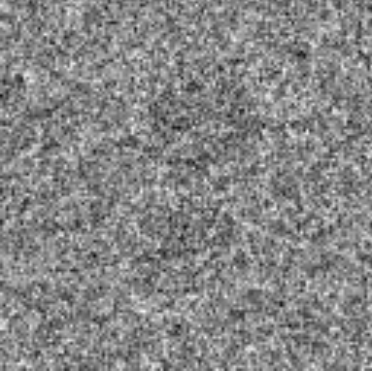}
	\includegraphics[width=0.32\textwidth]{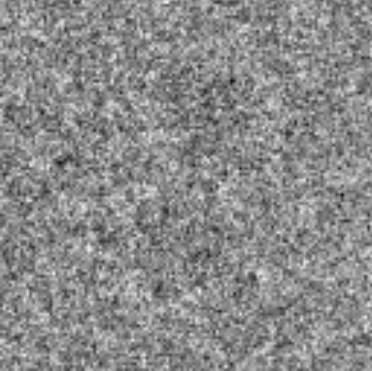}
	\includegraphics[width=0.32\textwidth]{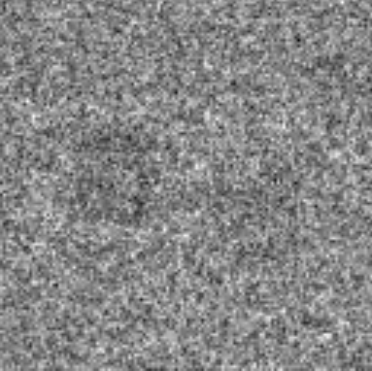}\\
	\includegraphics[width=0.32\textwidth]{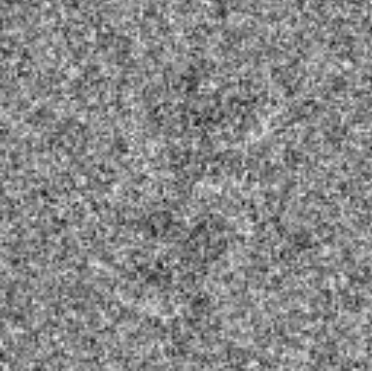}
	\includegraphics[width=0.32\textwidth]{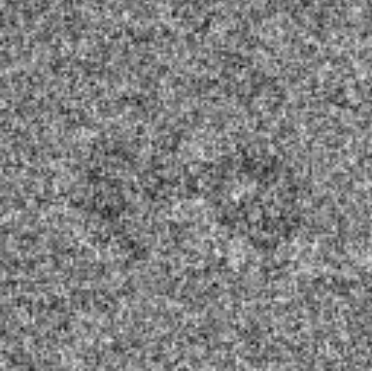}
	\includegraphics[width=0.32\textwidth]{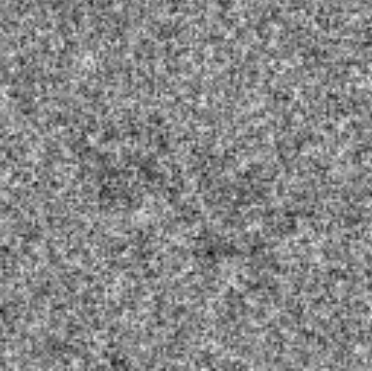}
	\end{subfigure}%
	&
	\includegraphics[width=0.17\textwidth,clip,trim=150 80 150 80]{figs/fig/thermus/sagd_is/3d_1}%
	\includegraphics[width=0.17\textwidth,clip,trim=150 80 150 80]{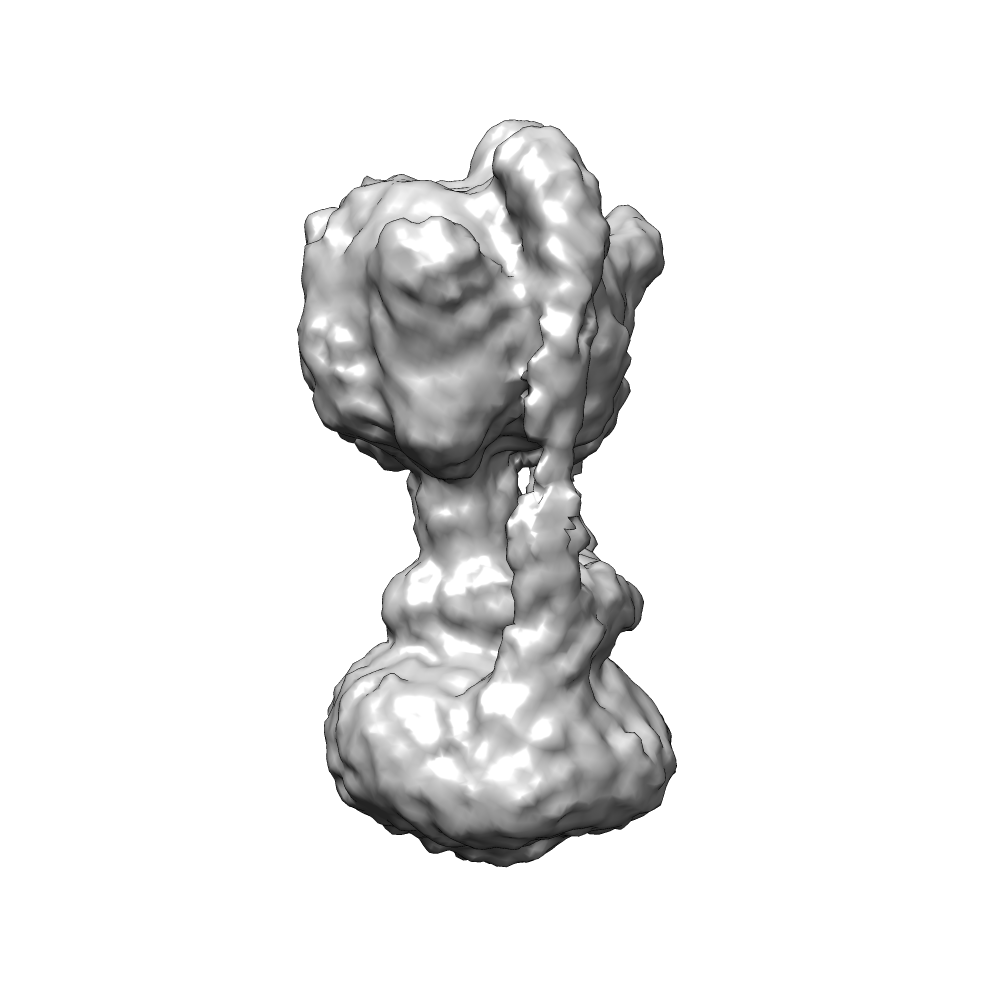}%
	&
	\includegraphics[width=0.17\textwidth,clip,trim=150 80 150 80]{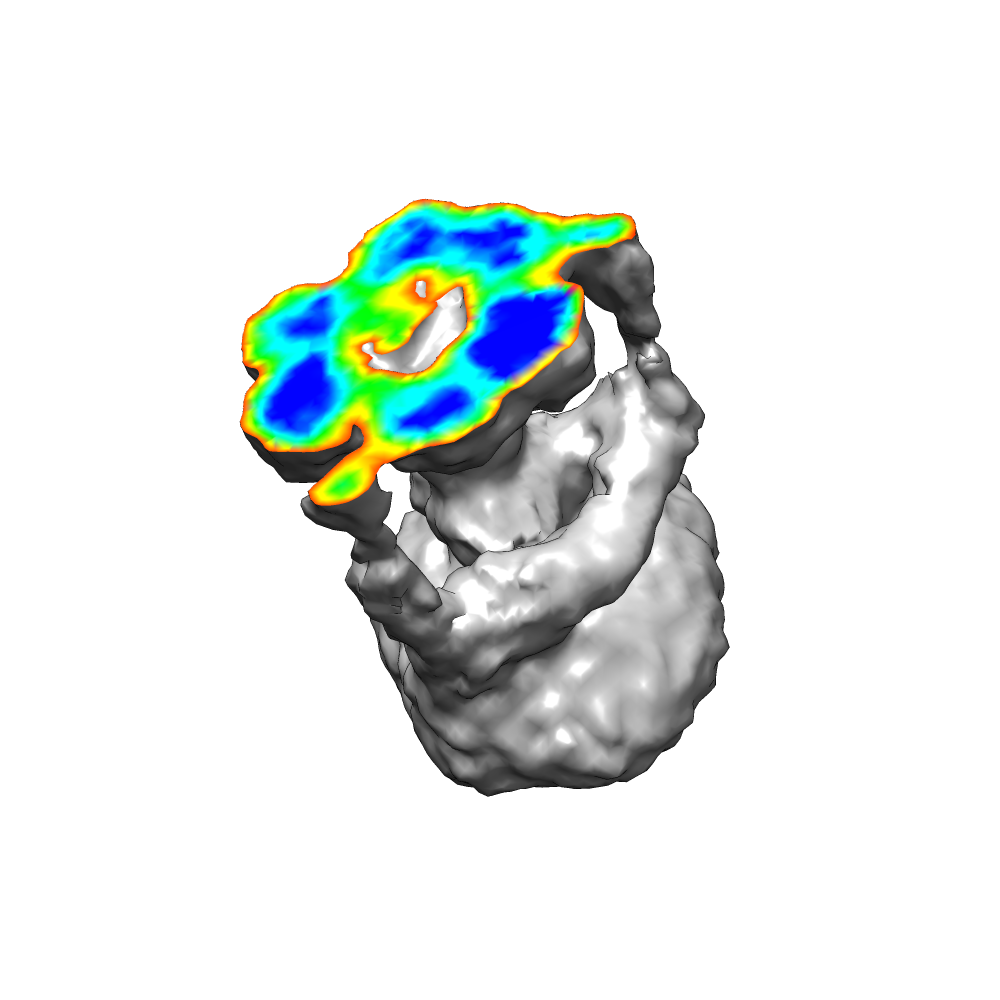}%
	\includegraphics[width=0.17\textwidth,clip,trim=150 80 150 80]{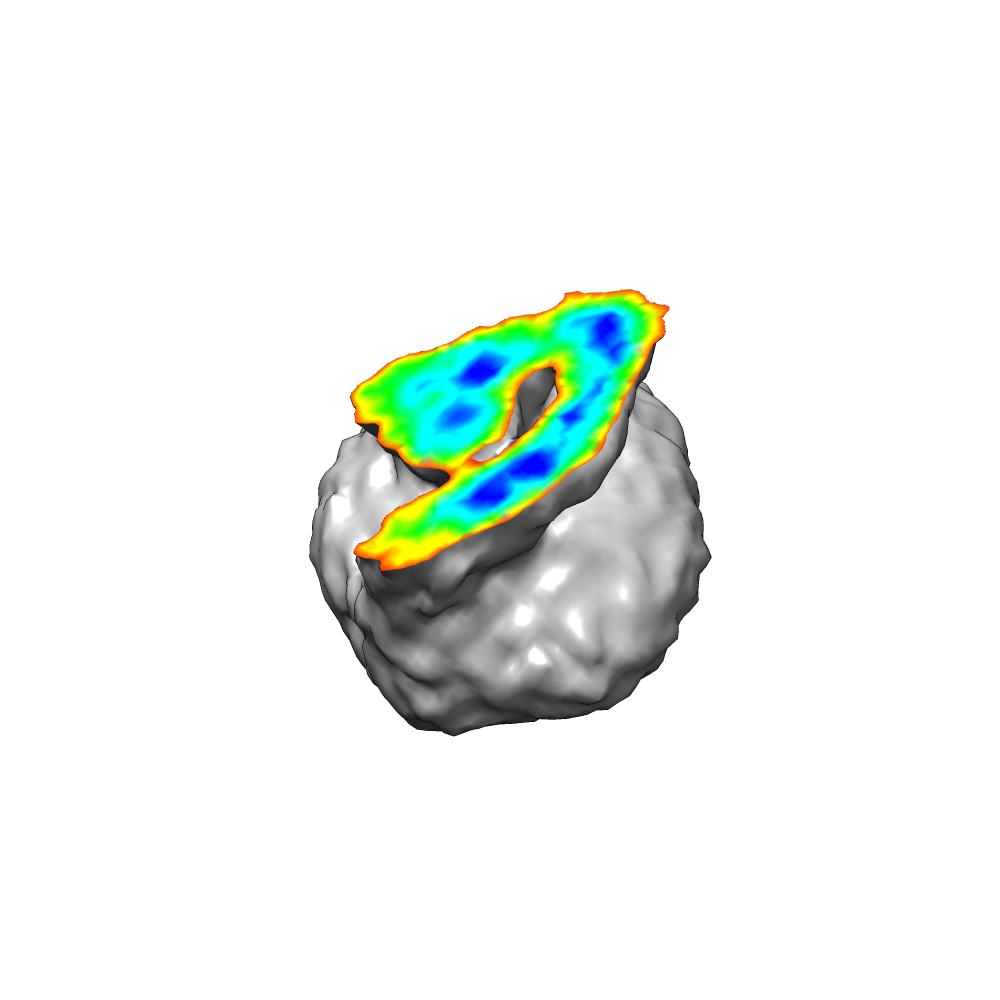}\\%
	\hline
	\dataname{\ \ \ \ \ \ \ \ \ Bovine ATPase} &
	\begin{subfigure}[b]{0.29\textwidth}
	\includegraphics[width=0.32\textwidth]{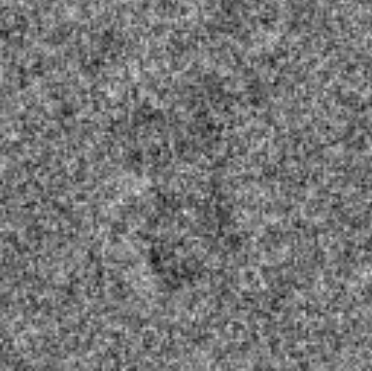}
	\includegraphics[width=0.32\textwidth]{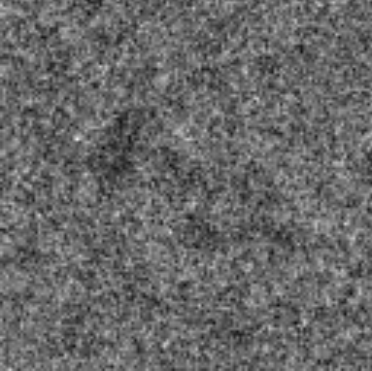}
	\includegraphics[width=0.32\textwidth]{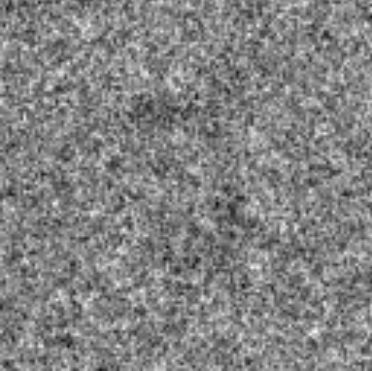}\\
	\includegraphics[width=0.32\textwidth]{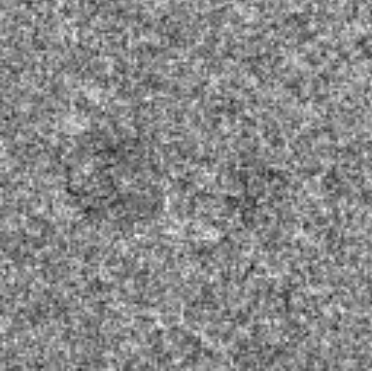}
	\includegraphics[width=0.32\textwidth]{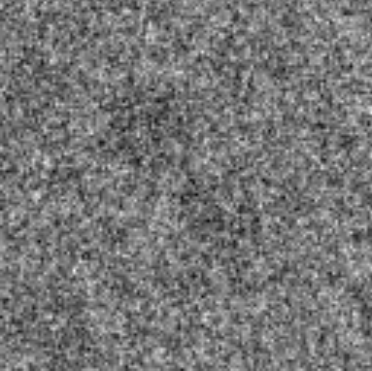}
	\includegraphics[width=0.32\textwidth]{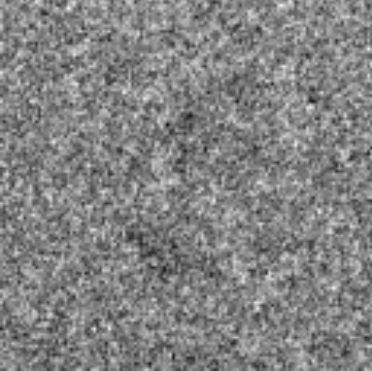}
	\end{subfigure}%
	&
	\includegraphics[width=0.17\textwidth,clip,trim=150 80 150 80]{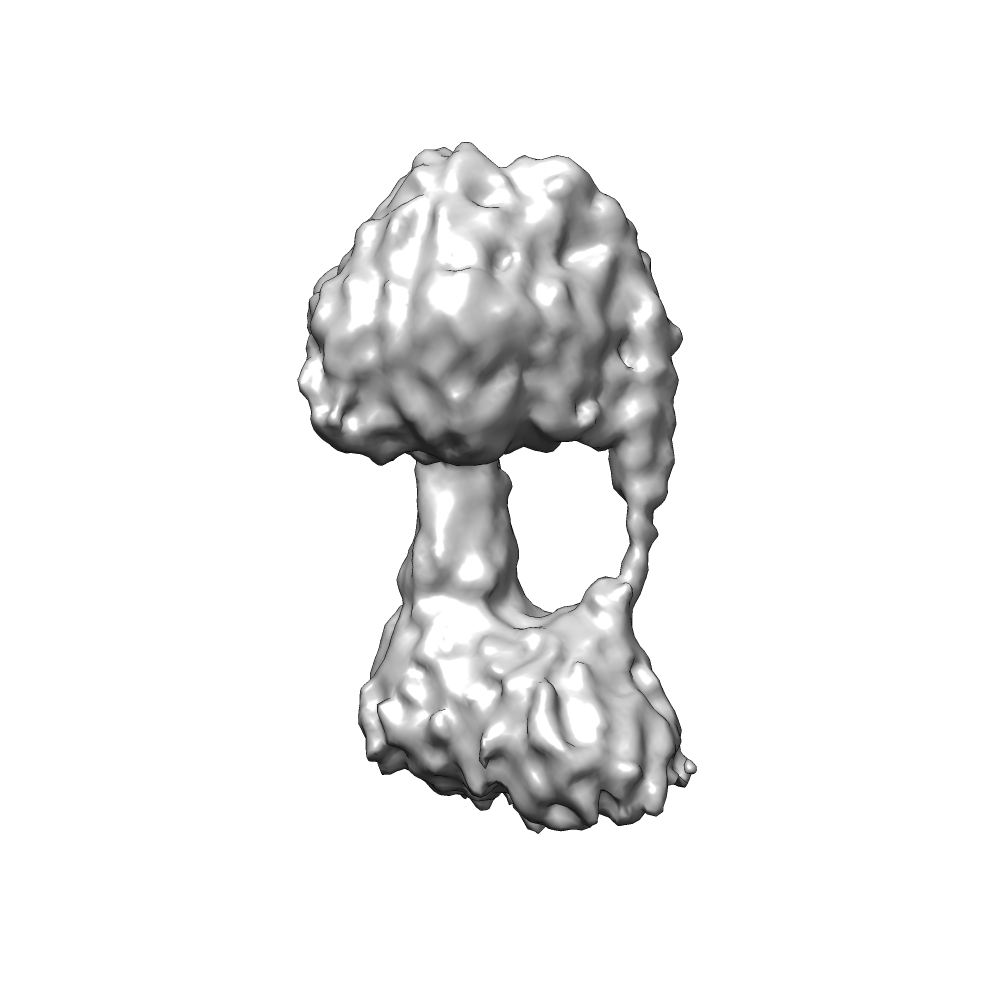}%
	\includegraphics[width=0.17\textwidth,clip,trim=150 80 150 80]{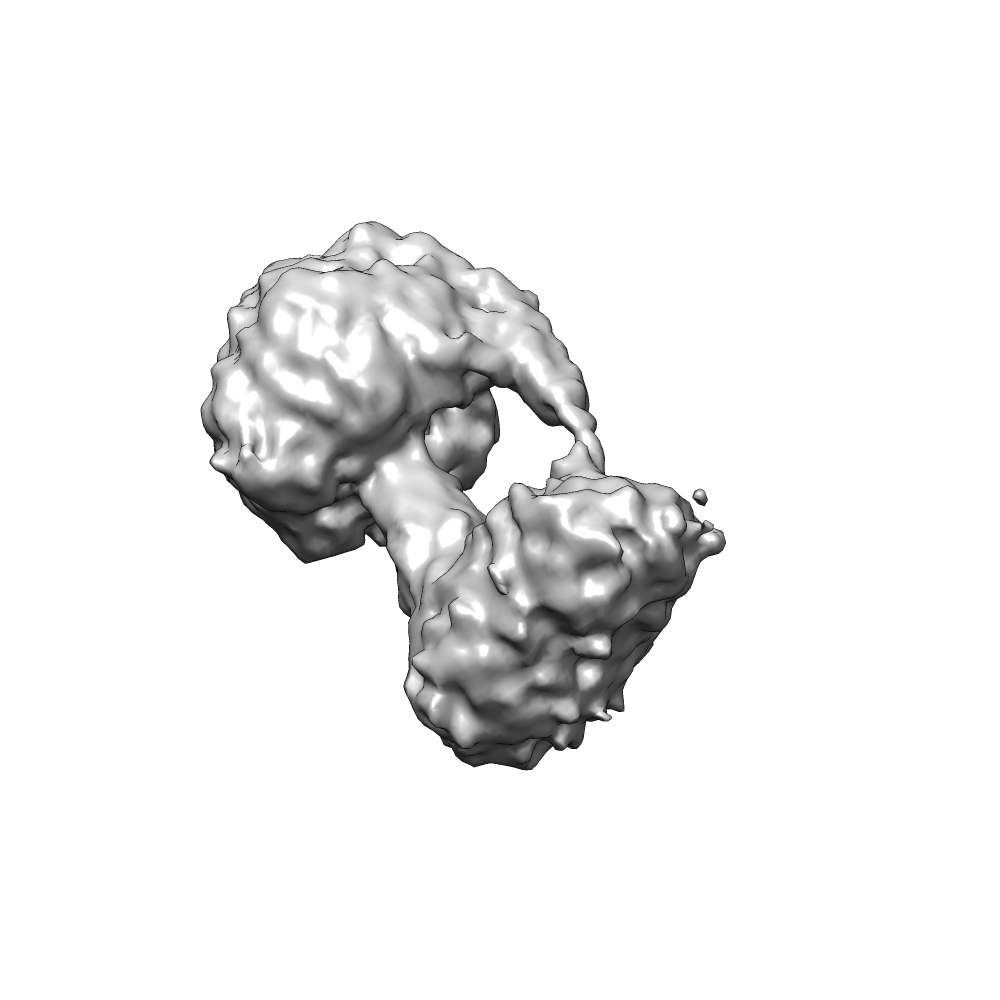}%
	&
	\includegraphics[width=0.17\textwidth,clip,trim=150 80 150 80]{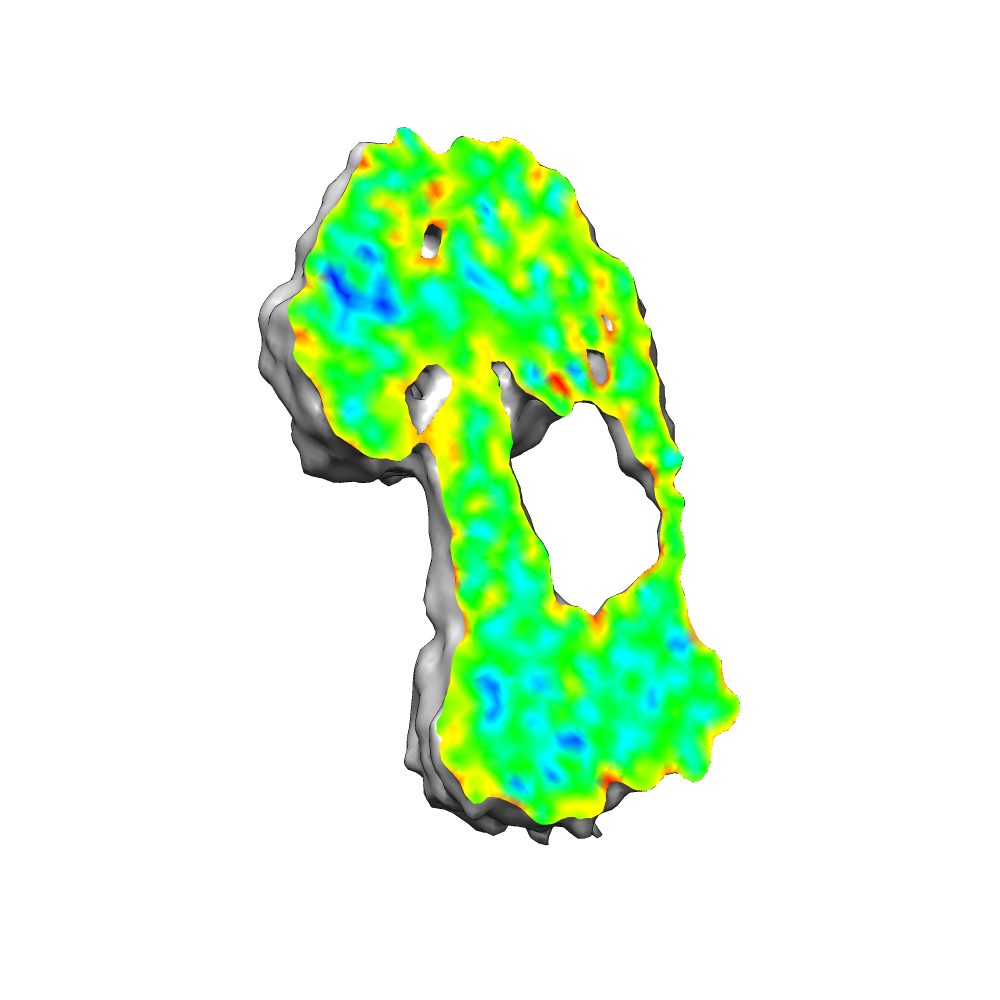}%
	\includegraphics[width=0.17\textwidth,clip,trim=150 80 150 80]{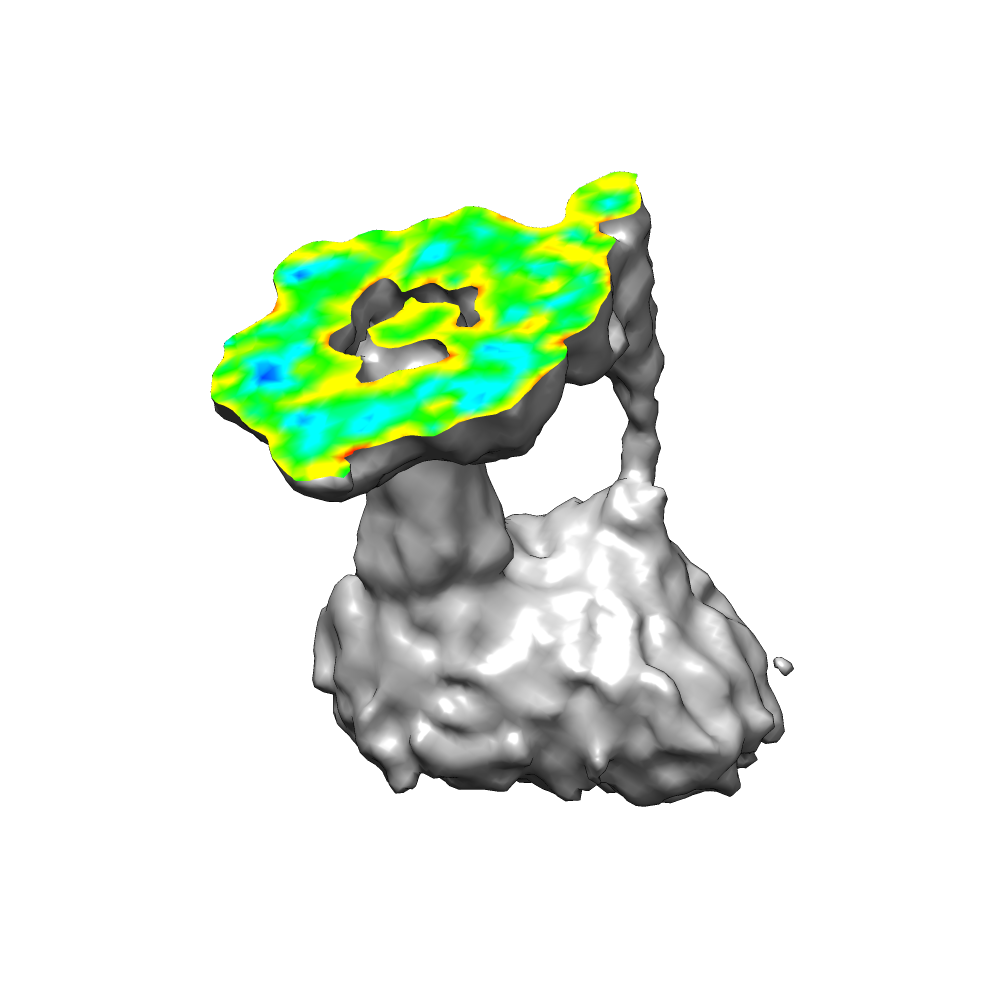}\\%
	\end{tabular}
\caption{\label{fig:resgrid}
Sample particle images (left), an isosurface of the reconstructed 3D density (middle)
and slices through the 3D density with colour indicating relative density (right) for
GroEL-GroES (top), thermus thermophilus ATPase (middle) and bovine mitochondrial ATPase
(bottom).
The relative root expected mean squared error (RREMSE) on a held-out test set was 0.99, 0.96
and 0.98 with values relative to the estimated noise level.
See Supplemental Material for more on the error measure.
Reconstructions took a day or less on a 16 core workstation.
\vspace*{-0.25cm}
}
\end{figure*}

\paragraph{Comparing Priors}
The above results used an exponential prior for the density at the voxels
of $\density_i$, however the presented framework allows for any continuous
and differentiable prior to be used.
To demonstrate this, we explored two other priors: an improper (uniform)
prior, $p(\density_i) \propto 1$, and a conditionally autoregressive (CAR) 
prior \citep{Besag1975}
$p(\density_i | \density_{-i}) = \mathcal{N}(\density_i|\frac{1}{26}\sum_{j\in\mbox{Nbhd}(i)} \density_j, \sigma_{CAR}^2)$
which is a smoothness prior biasing each voxel towards the mean of its 26
immediate neighbours $\mbox{Nbhd}(i)$.
Slices through the resulting densities on \emph{thermus} under these 
priors are shown in Figure \ref{fig:priors}.
With an improper uniform prior (Fig.\ \ref{fig:priors},left), there is 
significant noise visible in the background.
This noise is somewhat suppressed with the CAR prior 
(Fig.\ \ref{fig:priors},middle) however the best results are clearly
obtained using the exponential prior which suppresses the background noise 
without smoothing out internal details.

\section{Conclusions}

This paper introduces a framework for efficient 3D molecular
reconstruction from Cryo-EM images.
It comprises MAP estimation of 3D structure with a generative model, 
marginalization over 3D particle poses, and optimization using 
SAGD.  A novel importance sampling scheme was used to reduce the 
computational cost of marginalization.  The resulting approach
can be applied to large stacks of Cryo-EM images, providing 
high resolution reconstructions in a day on a 16-core workstation.

The problem of density estimation for Cryo-EM is a fascinating vision
problem. 
The low SNR in particle images makes it remarkable that any
molecular structure can be estimated, let alone the high resolution
densities which are now common.
Recent research \citep{Li2013} suggests that the combination of new 
techniques and new sensors may facilitate atomic resolution reconstructions
for arbitrary molecules.
This development will be ground-breaking in both biological and
medical research.

Beyond the work described in this paper, there remain 
a number of unresolved questions for future research. 
While an exponential prior was found to be effective, more 
sophisticated priors could be learned, potentially enabling higher 
resolution estimation without the need to collect more data and
providing a kind of of atomic-scale super-resolution.
The optimization problem is challenging, and, while SAGD
was successful here, it is likely that more efficient stochastic
optimization methods are possible by exploiting the problem structure
to a greater degree.
In order to encourage others to work on this problem, source
code will be available from the \href{http://www.cs.toronto.edu/\~mbrubake}{authors' website}.

\subparagraph*{Acknowledgements}
This work was supported in part by NSERC Canada and the CIFAR NCAP Program.
MAB was funded in part by an NSERC Postdoctoral Fellowship.
The authors would like thank John L.\ Rubinstein for providing data
and invaluable feedback.

{\small 
\bibliographystyle{IEEEtranSN}
\bibliography{refs}
}

\appendix

\section{Stochastic Optimization}
This section provides algorithmic details of the Stochastic Averaged Gradient Descent 
(SAGD) optimization method used for MAP estimation.
See the original SAGD paper \citep{LeRoux2012} for details.
Consider the objective function 
specified in Equation (\ref{eq:OptimProblem}), rewritten as a sum of functions over
subsets of the data:
\begin{align*}
	\obj(\density) &= -\log p(\density) - \sum_{i=1}^K\log p(\ftimg_i | \ctfparam_i, \ftdensity) \\
	&= \sum_{i=1}^K\left[-\frac{1}{K}\log p(\density) - \log p(\ftimg_i | \ctfparam_i, \ftdensity) \right] \\
	&= \sum_{i=1}^K \obj_i(\density) \\
\end{align*}
At each iteration $\iter$, SAGD computes the update given by
\begin{align*}
	\density_{\iter+1} &= \density_{\iter} - \frac{\epsilon}{L} \sum_{j=1}^K \left[ \densitygrad_j^\iter - \frac{1}{K} \frac{\partial}{\partial \density}\log p(\density)\right]
\end{align*}
where $\densitygrad_k^\iter$ is defined according to Equation (\ref{eq:sagd_grad}). In practice, the sum in the above update equation is not computed at each iteration, but rather a running total is maintained and updated as follows:
\begin{align*}
	\hat{\dobj}_{\iter} &= \sum_{k=1}^K \densitygrad_k^\iter \\
	\hat{\dobj}_{\iter+1} &= \hat{\dobj}_{\iter} - \densitygrad_{k_\iter}^{\iter} + \dobj_{k_\iter}(\density_{\iter}) \\
\end{align*}

The SAGD algorithm requires a Lipschitz constant $L$ which is not generally know.
Instead it is estimated using a line search algorithm where an initial value of $L$ is increased until the instantiated Lipschitz condition
$\obj(\density) - \obj(\density - L^{-1}\densitygrad) < \frac{\Vert \densitygrad \Vert^2}{2L}$
is met.
The line search for the Lipschitz constant $L$ is only performed once every 20 iterations.
Note that a more sophisticated line search could be performed if desired.
A good initial value of $L$ is found using a bisection search where the upper bound is the smallest $L$ found so far to satisfy the condition and the lower bound is the largest $L$ found so far which fails the condition.
In between line searches, $L$ is gradually decreased to try to take larger steps.
The entire SAGD algorithm is provided in Algorithm (\ref{alg:SAGD}).

\begin{algorithm}
\small
	\begin{algorithmic}
		\State Initialize $\density$ and $L$
		\State Initialize $\hat{\dobj} \gets 0$
		\State Initialize $\densitygrad_{k} \gets 0$ for all $k=1..K$
		\For {$\iter =1 .. \iter_{\text{max}}$}
			\State Select data subset $k_\iter$
			\State Compute objective gradient $\dobj_{k_\iter}(\density)$
			\State $\hat{\dobj} \gets \hat{\dobj} - \densitygrad_{k_\iter} + \dobj_{k_\iter}(\density)$
			\State $\densitygrad_{k_\iter} \gets \dobj_{k_\iter}(\density)$
			\State $\density \gets \density - \frac{\epsilon}{L} \left[\hat{\dobj} - \frac{\partial}{\partial \density}\log p(\density)\right]$
			\If {mod($\iter$,20) == 0}
			    \State \emph{Perform line search}
			    \While { $\obj_{k_\iter}(\density) - \obj_{k_\iter}(\density - L^{-1}\densitygrad_{k_\iter}) < \frac{\Vert \densitygrad_{k_\iter} \Vert^2}{2L}$ }
			        \State $L \gets 2L$
			    \EndWhile
			\Else
			    \State $L \gets \frac{K}{2^{\frac{1}{150}}}$
			\EndIf
		\EndFor
	\end{algorithmic}
	\caption{SAGD}
	\label{alg:SAGD}
\end{algorithm}

\section{Importance Sampling}

Importance Sampling is a key part of the proposed reconstruction method for Cryo-EM 
and provides large speedups during optimization. We use importance sampling to efficiently
compute the discrete sum in Equation (\ref{eq:ApproxMarginalization}). 
Note that importance sampling is applied independently for each image in the dataset,
since the orientations and shifts which correspond to important terms in the discrete sum
can be different for each image.

In practice, we split the outer sum in Equation (\ref{eq:ApproxMarginalization}) into a double summation, 
one over orientations on the sphere and one over in-plane rotations of images and projections.
We then compute each of the three sums (over shift, in-plane rotation, and orientation) with 
and independent importance sampler. This is equivalent to computing the full sum in 
Equation (\ref{eq:ApproxMarginalization}) using a single importance sampler with an importance
distribution that is factored into three parts, one for each of shift, in-plane rotation, 
and orientation. 
This factoring is necessary, as the memory requirements of storing a 
fully joint importance distribution for each image in the dataset would become infeasible for 
high-resolution reconstructions.

For each of the three importance samplers, the importance distribution at each iteration is 
constructed according to Equation (\ref{eq:impsamp_mixing}). At the first iteration during which 
a particular image is seen, the importance distribution is simply uniform, and in fact we explicitly 
sample every point once. The $\phi$ values resulting from this computation are stored. 
At the next iteration during which the same image is seen, these $\phi$ values are used in 
Equation (\ref{eq:impsamp_mixing}) to construct a non-uniform importance distribution which is then
sampled from. We use a number of samples proportional to the effective sample size of the importance 
distribution, so the number of samples used naturally decreases as the importance distribution becomes
more peaked, leading to large speedups at late iterations during optimization.

\begin{algorithm}
\small
	\begin{algorithmic}
		\State Given $\phi_i$ for $i \in \mathfrak{I}$ from previous iteration
		\For {$j \in 1..J$}
			\For {$i \in \mathfrak{I}$}
				\State Compute $\kernmat_{i,j}$
			\EndFor
		\EndFor
		\State $\hat{\phi}_j \gets \sum_{i\in\mathfrak{I}} \phi_{i}^{1/T} \kernmat_{i,j}$ $\;\;\forall j\in 1..J$
		\State $Z \gets \sum_j \hat{\phi}_j$
		\State $q_j \gets (1-\alpha) Z^{-1} \hat{\phi}_j + \alpha \psi_j$ $\;\;\forall j\in 1..J$
		\State $s \gets \left( \sum_j q_j^2 \right)^{-1}$
		\State $N \gets s_0s$
		\State $\mathfrak{I} \gets \varnothing$
		\For {$k \in 1..N$}
			\State $i \gets $ sample from $q$
			\State insert $i$ into $\mathfrak{I}$
		\EndFor
		\State Use $\mathfrak{I}$ to compute $\phi_i$ for next iteration
	\end{algorithmic}
	\caption{Importance Sampling}
	\label{alg:IMP}
\end{algorithm}

In Equation (\ref{eq:impsamp_mixing}), the previous $\phi$ values are not directly used, but rather they are annealed by a temperature parameter and then smoothed by a kernel matrix. Both of these 
steps serve to guard against importance distributions which are too peaked around large $\phi$ values, 
which would inhibit the importance sampler from exploring the domain. The kernel matrix also serves the purpose of allowing use of $\phi$ values from a previous iteration even if the resolution of quadrature
points being used has increased at the current iteration. The Von Mises-Fisher kernel is used for 
orientations and in-plane rotations, while a Gaussian kernel is used for shift:
\begin{align*}
	\kernmat_V(d_i, d_j;\kappa_V) &\propto \exp(\kappa_V d_i^T d_j) \\
	\kernmat_G(t_i, t_j;\kappa_G) &\propto \exp(-\kappa_G \|\shiftdir_i - \shiftdir_j\|^2) 
\end{align*}
where $\kappa_V$ and $\kappa_G$ are precision parameters for each kernel
which are set based on the resolution of the quadrature scheme used at
the previous $\phi$ values, $d_i$ and $d_j$ are the quadrature directions (in $\mathcal{S}^2$ for particle orientation and $\mathcal{S}^1$ for in-plane rotation, and $\shiftdir_i$ and $\shiftdir_j$ are the quadrature shift values (in $\R^2$).

The algorithm for constructing an importance distribution and sampling from it
are given in Algorithm (\ref{alg:IMP}).
The sampled values are then used to compute (\ref{eq:ISApprox}).
Note that some quadrature points can end up being sampled multiple times,
this is detected and the value reused to reduce computation.

\section{Error Measures}
Because ground-truth is rarely available for Cryo-EM, measuring accuracy is
often difficult.
Traditionally, the field has used the \emph{Fourier Shell Correlation} (FSC)
to measure the resolution of a solved structure.
The so-called gold-standard FSC works by splitting the dataset in half,
estimating two densities separately and the computing the normalized correlation in 
Fourier space as a function of frequency.
This curve would then be thresholded to provide an estimate of accuracy.
However, we note that this measure is actually estimating the variance of the 
estimator, not the accuracy of the density it has produced.
Further it is only theoretically justifiable when the estimator is unbiased, 
which is not true of the method proposed here or with other likelihood-based
Bayesian methods such as RELION.

Instead, we introduce a novel metric based on reconstruction error of a held
test set.
To quantify the ability of marginal likelihood methods, such as ours, to model
and explain the observed data we introduce the \emph{Expected Mean Squared Error} 
\begin{equation}
\mathcal{E}^2(\img | \ctfparam, \density) \equiv
E_{\projdir,\shiftdir | \img, \ctfparam, \density}\left[ \Vert \img - \ctf{\ctfparam} \shift{\shiftdir} \proj{\projdir} \density \Vert^2 \right]\\
\end{equation}
to be the expectation of the squared error between the image and its 
reconstruction under the image formation model.
Note that the expectation is conditioned on the current density and the CTF
parameters and is taken over the unknown pose and translation,
$\projdir$ and $\shiftdir$.
After switching to Fourier space and with some manipulation
$\mathcal{E}^2(\img | \ctfparam, \density)$ becomes
\begin{equation}
Z^{-1} \int_{\R^2} \int_{\SO} \Vert \ftimg - \ftctf{\ctfparam} \ftshift{\shiftdir} \ftproj{\projdir} \ftdensity \Vert^2 p(\ftimg | \ctfparam, \projdir, \shiftdir, \ftdensity ) p(\projdir) p(\shiftdir) d\projdir d\shiftdir
\end{equation}
where the 
\begin{equation}
Z = \int_{\R^2} \int_{\SO} p(\ftimg | \ctfparam, \projdir, \shiftdir, \ftdensity ) p(\projdir) p(\shiftdir) d\projdir d\shiftdir
\end{equation}
is a normalization constant.
Computing this would be computationally expensive, instead we use an
importance sampling based approximation, $\hat{\mathcal{E}}^2(\img | \ctfparam, \density)$,
\begin{equation}
\hat{Z}^{-1}
\sum_{j \in \mathfrak{I}^{\projdir}}
\sum_{\ell \in \mathfrak{I}^{\shiftdir}}
\frac{w_j^{\projdir} w_\ell^{\shiftdir}}{N_{\projdir} \isp_j^{\projdir} N_{\shiftdir} \isp_\ell^{\shiftdir}}
p_{j,\ell} \Vert \ftimg - \ftctf{\ctfparam} \ftshift{\shiftdir} \ftproj{\projdir} \ftdensity \Vert^2
\end{equation}
where
\begin{equation}
\hat{Z} =
\sum_{j \in \mathfrak{I}^{\projdir}}
\sum_{\ell \in \mathfrak{I}^{\shiftdir}}
\frac{w_j^{\projdir} w_\ell^{\shiftdir}}{N_{\projdir} \isp_j^{\projdir} N_{\shiftdir} \isp_\ell^{\shiftdir}}
p_{j,\ell}
\end{equation}
is the approximation of the normalization constant.
The above quantities can be readily computed along with the main likelihood
computation using the same importance sampling scheme described above.

We compute the average value of $\hat{\mathcal{E}}^2(\img | \ctfparam, \density)$
on a held out set of test images whose gradients are never used.
To normalize for different datasets we report the \emph{Relative Root Expected
Mean Squared Error} (RREMSE) as
\begin{equation}
\sqrt{ \frac{1}{\sigma^2 N_{\mbox{test}}}\sum_{\img} \hat{\mathcal{E}}^2(\img | \ctfparam, \density)}
\end{equation}
where the sum is taken over the test set which has $N_{\mbox{test}}$ images
and $\sigma^2$ is the noise variance of the dataset.
Values near 1 indicate that the data is being well explained.

\end{document}